\documentclass[a4paper,fleqn]{cas-dc}

\usepackage[authoryear,longnamesfirst]{natbib}
\usepackage{multirow}
\usepackage{moreverb,url}

\def\tsc#1{\csdef{#1}{\textsc{\lowercase{#1}}\xspace}}
\tsc{WGM}
\tsc{QE}

\begin{document}
\let\WriteBookmarks\relax
\def\floatpagepagefraction{1}
\def\textpagefraction{.001}

% Short title
\shorttitle{A Multi-granularity Interaction Network for Relational Triple Extraction}

% Short author
\shortauthors{Luo, Chen and Zhan}

% Main title of the paper
\title[mode = title]{A Multi-granularity Interaction Network for Relational Triple Extraction}

% \tnotemark[1]
% \tnotetext[1]{}

% First author
\author[suda]{Xiaocheng Luo}
\ead{xcluo@stu.suda.edu.cn}

\author[gz]{Yanping Chen}
\ead{ypench@gmail.com}
\cormark[1]
\cortext[1]{Corresponding author}

\author[ece]{Zaifu Zhan}
\ead{zhan8023@umn.edu}

\affiliation[suda]{organization={School of Computer Science and Technology},%
    addressline={Soochow University}, 
    city={Suzhou},
    postcode={215006}, 
    state={Jiangsu},
    country={PR China}}

\affiliation[gz]{organization={State Key Laboratory of Public Big Data, College of Computer Science and Technology},
	addressline={Guizhou University}, 
	city={Guiyang},
	postcode={550025}, 
	state={Guizhou},
	country={PR China}}

\affiliation[ece]{organization={Department of Electrical and Computer Engineering},%
    addressline={University of Minnesota}, 
    city={Minneapolis},
    postcode={55455},
    state={MN},
    country={USA}}

\begin{abstract}
Relational triple extraction identifies entity relations from raw sentences without depending on entity type information. The performance of traditional methods is seriously degenerated by ``unseen'' entities, because these methods can't build contextual dependencies relevant ``unseen'' entities. This paper addresses this challenge by proposing a multi-granularity interaction network that leverages word-word, entity-relation, and head-tail interactions to capture deep semantic dependencies. By incorporating multiple iterations, the multi-granularity interaction network is effective to refine the representations corresponding to head and tail entities. It enhances the global semantic features associated with relational triples, aiding in the remediation of triples that may have been overlooked or omitted in earlier steps. Strong and consistent performance is achieved by our method on two rearranged datasets, with a significant improvement in identifying relational triples, including those involving ``unseen'' named entities. For example, our model improves F1-score by 3.1\%/20.5\% over previous approaches on the NYT-R/WebNLG-R datasets in the Unseen metric. Code: \href{https://github.com/xiaocen-luo/Interaction-Network}{https://github.com/xiaocen-luo/Interaction-Network.}
\end{abstract}

% Research highlights（原文无，需作者补写3-5条，每条不超过85字符）
\begin{highlights}
\item A multi-granularity interaction network is proposed for relational triple extraction.
\item Word-word, entity-relation and head-tail interactions are jointly leveraged.
\item The model handles entirely seen, partially seen and unseen relational triples.
\item Consistent improvements are observed on NYT-R and WebNLG-R datasets.
\end{highlights}


\maketitle

\section{Introduction}
\label{sec:Introduction}

The task of Relational Triple Extraction (RTE) involves both extracting entity pairs and recognizing their relation types from a raw sentence. The output can be represented in the form of a 3-tuple ($h$, $r$, $t$) denoted to three parameters: a $head$ entity, a $relation$ type and a $tail$ entity. Because relation triple extraction ignores semantic types of named entities, it is effective to retrieve semantic relations between all possible named entities and avoid the error propagation problem caused by classifying the semantic types of named entities. In recent years, this task has received a lot of attention in the field of natural language processing. Knowledge graph construction \citep{dong2014knowledge}, information retrieval \citep{sinha2015overview}, and text semantic analysis \citep{hua2015short} are among the downstream tasks that commonly benefit from this technique.

Large Language Models (LLMs) \citep{achiam2023gpt, team2024qwen2, dubey2024llama} have recently achieved impressive results in many natural language processing applications. For instance, they work well for Question Answering \citep{yona2024narrowing} and Code Generation \citep{li2022competition}. However, these same models tend to experience a notable performance decline when tackling information extraction tasks. For instance, as reported by \cite{ding2024improving}, LLMs are able to perform triple extraction from basic sentences. This can be achieved through few-shot learning or fine-tuning, assuming appropriate instructions are provided. Despite this capability, they frequently suffer from poor performance in NLP tasks relevant to complex sentence structures. The performance drop stems from the fact that sentences often contain multiple entities simultaneously, particularly in the relational triple extraction. Within the scope of relational triple extraction, all entities in a given sentence share overlapping contextual cues, leading to intricate semantic structures. Standard practice for generative models is to convert a sentence into a dense vector, which hampers their ability to tell entities apart while decoding overlapping relational triples.

Early works on relational triple extraction are usually based on a pipelined framework \citep{zelenko2003kernel, chan2011exploiting}, which decomposes the task into a two-step process involving entity recognition followed by relation classification. Because each incorrect entity will be paired with all entities for relation classification, the error propagation problem is very influential in the pipeline framework. To address this problem, joint methods are widely adopted to extract entity pairs together with their relations via a unified approach \citep{zheng2021prgc, wei2020novel, li2021tdeer, an2024rtf}. These methods demonstrate strong promise for tackling the relational triple extraction problem.

In joint relational triple extraction, every relation triple is directly extracted without depends on any prior information about named entities. Because the task imposes no limits about the types of named entities, all named entities can be used as candidates of a relation triple. Therefore, to support this task, it is critical to identify relation triples composed of ``unseen'' named entities in the training process. According to the ``visibility'' of triple parameters, relational triples in the testing dataset can be divided into three categories \citep{lee2022does}: $\emph{entirely seen}$ (Entire), $\emph{partially seen}$ (Partial), $\emph{unseen}$ (Unseen). The $\emph{entirely seen}$ refers to relational triples with the same two entity parameters ($head$ and $tail$) that occurred within the training set as well as the testing set. In a $\emph{partially seen}$ triple, either a tail entity or a head entity co-occurred in training and testing datasets. Table \ref{table:dataset info} lists the statistical information about unseen relational triples in four evaluation datasets including $\mathrm{NYT}$, $\mathrm{WebNLG}$, along with their modified versions $\mathrm{NYT}$-$\mathrm{R}$ and $\mathrm{WebNLG}$-$\mathrm{R}$.

\begin{table}[h]
    \footnotesize
	\caption{ \label{table:dataset info} The number (ratio) of triple types in the testing datasets.}
    \begin{tabular}{lccc}
        \hline
        Dataset & Entire & Partial & Unseen  \\
        \hline
        $\mathrm{WebNLG}$ & 756 ($\textcolor{red}{\mathbf{85.23}\%}$) & 110 (12.40$\%$) & 21 (2.37$\%$)  \\
        $\mathrm{NYT}$ & 2672 ($\textcolor{red}{\mathbf{77.07\%}}$) & 663 (19.12$\%$) & 132 (3.81$\%$)  \\
        $\mathrm{WebNLG}$-$\mathrm{R}$ & 439 (44.17$\%$) & 338 (34.00$\%$) & 217 (21.83$\%$)   \\
        $\mathrm{NYT}$-$\mathrm{R}$ & 823(13.81$\%$) & 3985 (66.89$\%$) & 1150 (19.30$\%$) \\
        \hline
    \end{tabular}
\end{table}

In the above table, $\mathrm{NYT}$ and $\mathrm{WebNLG}$ are two public evaluations for relational triple extraction, where `Entire' accounts for $\mathbf{77.07\%}$ and $\mathbf{85.23\%}$ on the $\mathrm{NYT}$ and $\mathrm{WebNLG}$ datasets, respectively. As shown in this table, the two datasets only contain a small number of unseen relational triples. To support practical applications of relational triple extraction, \cite{lee2022does} rearranged the two datasets to generate the $\mathrm{NYT}$-$\mathrm{R}$ and $\mathrm{WebNLG}$-$\mathrm{R}$ datasets. They contain a larger number of unseen triple, which makes the datasets closer to real applications.

To show the influence of unseen relational triples in real applications, an experiment was conducted on the four datasets $\mathrm{NYT}$, $\mathrm{WebNLG}$, $\mathrm{NYT}$-$\mathrm{R}$ and $\mathrm{WebNLG}$-$\mathrm{R}$. Three recent state-of-the-art models such as CasRel \citep{wei2020novel}, TPLinker \citep{wang2020tplinker}, and PRGC \citep{zheng2021prgc} are implemented on the four datasets. The result is shown in Table \ref{table:RTE on rearranged datasets}.

\begin{table}[!h]
    \small
	\caption{\label{table:RTE on rearranged datasets} The F1 performance achieved by the current best-performing model across four datasets \citep{lee2022does}.}
    \begin{tabular}{lccccccc}
        \hline 
        \multirow{2}{*}{Model} & \multicolumn{4}{c}{Dataset} \\
        \cline{2-5}
         & $\mathrm{WebNLG}$ & $\mathrm{WebNLG}$-$\mathrm{R}$ & $\mathrm{NYT}$ & $\mathrm{NYT}$-$\mathrm{R}$ \\ 
        \hline
        CasRel  & 88.3 & 68.6($\mathbf{\textcolor{red}{\downarrow 19.7}}$) & 90.1 & 62.9($\mathbf{\textcolor{red}{\downarrow 27.2}}$) \\
        TPLinker  & 89.0 & 69.1($\mathbf{\textcolor{red}{\downarrow 19.9}}$) & 92.4 & 64.7($\mathbf{\textcolor{red}{\downarrow 27.7}}$) \\
        PRGC  & 88.0 & 61.8($\mathbf{\textcolor{red}{\downarrow 26.2}}$) & 89.1 & 62.6($\mathbf{\textcolor{red}{\downarrow 26.5}}$) \\
        \hline
    \end{tabular}
\end{table}

The experiment showed that the performance is degenerated seriously when evaluated on the $\mathrm{WebNLG}$-$\mathrm{R}$ and $\mathrm{NYT}$-$\mathrm{R}$. This underscores that current methodologies excessively depend on the model's memory capabilities while neglecting its generalization abilities, which is inefficient in real applications.

\begin{figure}
	\centering
	\includegraphics[width=0.85\linewidth]{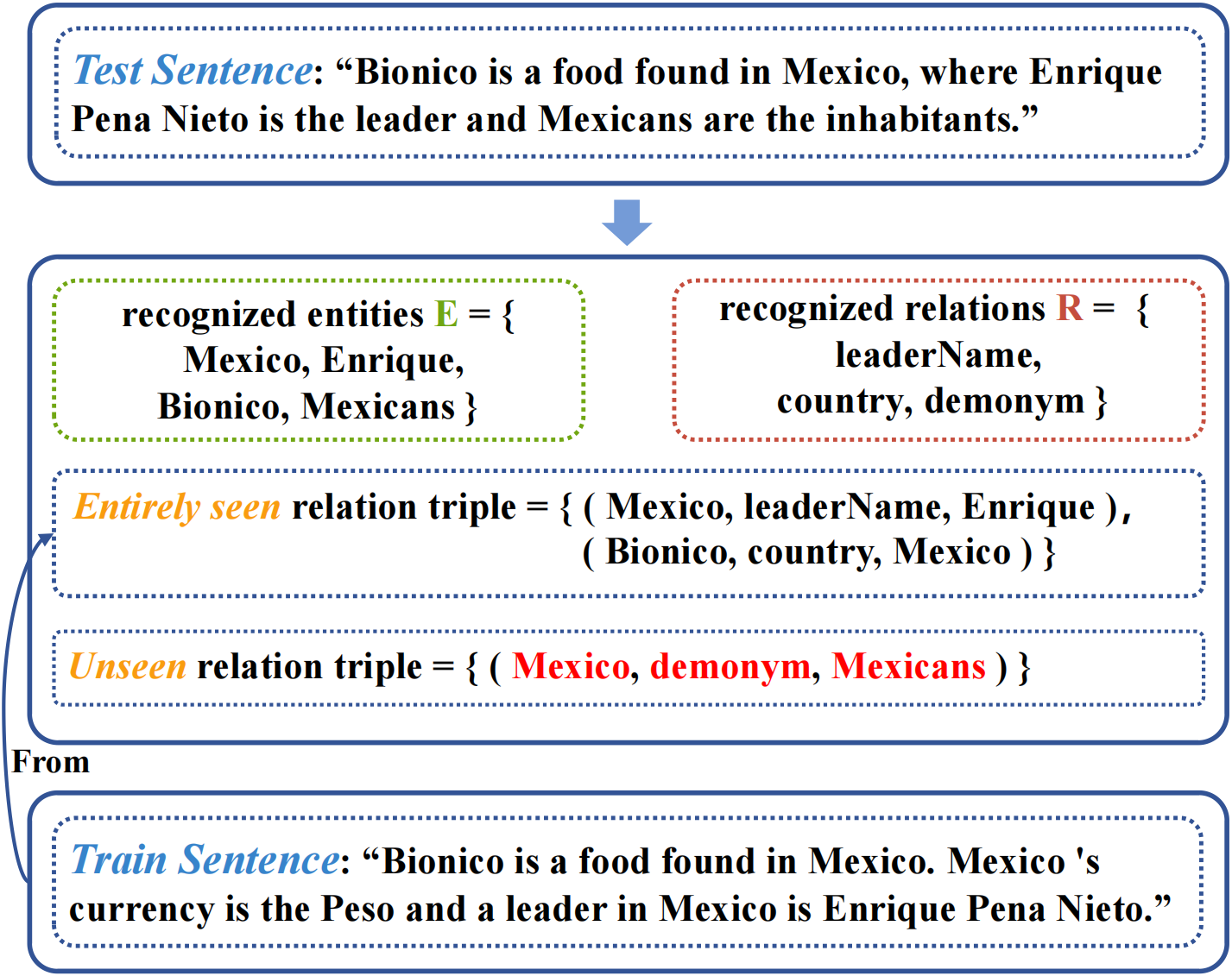}
	\caption{\label{fig:reasoning pattern} An example contains two types of relational triples.}
\end{figure}

Before giving the motivation of our method, an example is first given to show the characteristics of unseen relational triples. A testing sentence ``Bionico is a food found in Mexico, where Enrique Pena Nieto is the leader and Mexicans are the inhabitants.''. It contains four named entities (``Bionico'', ``Mexico'', ``Enrique'' and ``Mexicans'') and three relations (``leaderName'', ``country'' and ``demonym''). They generate three relational triples, which is illustrated in Figure \ref{fig:reasoning pattern}.

As shown in the bottom of Figure \ref{fig:reasoning pattern}, there is another sentence in the training dataset  ``Bionico is a food found in Mexico. Mexico's currency is the Peso and a leader in Mexico is Enrique Pena Nieto''. Therefore, the testing sentence contains two entirely seen tripes (Mexico, leaderName, Enrique) and (Bionico, country, Mexico), an unseen tripe (Mexico, demonym, Mexicans). Entirely seen relational triples tend to be easier to identify. On the other hand, the unseen triple is more challenging.

The motivation for developing multi-granularity interaction networks stems from several observations. First, entity pairs of unseen relational triples in the sentence can be inferred through interactions between head and tail. For example, in Figure \ref{fig:reasoning pattern}, the entity pair of the unseen relational triple makes a new relational connection through the head entity ``Mexico'' and the tail entity ``Mexicans''. Second, the relation types of unseen relational triples may be obtained through entity-relation interactions. For example, in Figure \ref{fig:reasoning pattern}, the specific logical structure ``one is the resident'' aids determine the type of the triple of relations as ``demonym'', which requires the interaction of a sufficient amount of semantic information. 

Motivated by the above observations, we introduce a multi-granularity interaction network designed to jointly extract relational triples across three categories: \emph{entirely seen}, \emph{partially seen}, and \emph{unseen}. ``multi-granularity'' refers to the model's ability to integrate different levels of interaction (or granularities) to improve relational triple extraction. Our model leverages the semantic interactions of word-word interaction, entity-relation interaction and head-tail interaction. The word-word interaction reinforces the semantic features of elements within a 2D sentence encoding, which makes the use of relation and entity information. The entity-relation interaction captures long-distance semantic dependencies between relations and entities, effectively filtering out the semantic features related to redundant relational triples. The head-tail interaction enhances the global semantic features associated with relational triples, aiding in the remediation of triples that may have been overlooked or omitted in earlier steps. By incorporating multiple iterations, the multi-granularity interaction network is effective to progressively refine the representations corresponding to head and tail entities. Evaluated on two rearranged datasets, our model consistently outperforms the compared baselines for relational triple extraction. Our work makes the following contributions:

\begin{itemize}
	\item[1)] A multi-granularity interaction network is proposed for extracting relational triples, that makes full use of word-word interactions, entity-relation associations, and head-tail interactions to extract entirely seen, partially seen, and unseen relational triples.
	\item[2)] Extensive experiments are performed on two rearranged datasets, where our method achieves the best results among all compared baselines. Further analytical results confirm the efficacy of our proposed approach for relational triple extraction.
\end{itemize}

\section{Related Work}
\label{sec:related_work}

Early studies  commonly adopted pipeline-based approaches to support relational triples extraction \citep{zelenko2003kernel, zhou2005exploring, chan2011exploiting}. Pipeline methods split the RTE problem into independent components: entity extraction followed by relation classification. These approaches have two main drawbacks. First, it ignores the possible interplay between entity identification and relation categorization. Second, it suffers from vulnerability to error propagation.

To overcome pipeline limitations, researchers now prioritize joint extraction methods \citep{sui2023joint, ning2023od, zhang2023rs}, which can extract entities and relations simultaneously. Related works on joint relational triple extraction fall into three types: tagging-based, Seq2Seq-based, and table-filling approaches. Each of these categories is outlined below.

\subsection{Tagging based Methods}

Tagging-based methods involve assigning tags to every token within a sentence for identifying entity boundaries and their relations. For instance, \cite{zhan2022simple} frames the relation extraction task as a multi-label problem that iteratively processes $n(n+1)/2$ tokens. For every distinct token pair, they incorporate bidirectional relations, utilizing dependency relations to capture directional information between entity pairs. \cite{Wang2023CLFMCL} introduced a relation-aware sequence labeling decoder, which utilizes a filter-attention mechanism to emphasize informative features. Subsequently, an additional matrix is employed to integrate entity pair information. \cite{he2023bitcoin} present a framework implementing taggers in two directions. It enables triples extraction from subject to object and vice versa. Tagging based methods have achieved good performance for identifying triples without overlapping. Because they usually rely on semantic expressions associated with a pair of named entities within a triple, performance seriously degrades when identifying relational triples with ``unseen'' entities.

\subsection{Seq2Seq based Methods}

Sequence-to-sequence approaches typically view a relational triple as a token sequence, thus casting the RTE problem as a generative task. The goal is to generate relational triples in an autoregressive manner. For example, \cite{zeng2018extracting} tackle the issue of overlapping relational triples by employing a copy mechanism. Meanwhile, \cite{zeng2019learning} incorporate reinforcement learning into their Seq2Seq framework and investigate the ordering issue in triple generation. Furthermore, \cite{zeng2020copymtl} design a multi-task learning architecture that utilizes a copy mechanism to address the difficulty posed by multi-token entities. \cite{nayak2020effective} present a new representation method that proves effective at extracting multiple triples containing overlapping entities, including those with multi-token entities within a sentence.
The main problem of the Seq2Seq based approach is that these methods should compress sentences into dense representations, which leads to serious information loss.

\subsection{Table Filling based Methods}

Approaches based on table filling convert the relational triple extraction problem into a table-filling procedure. \cite{wang2020tplinker} frames joint extraction as a problem of linking token pairs. They propose a handshaking labeling strategy that matches the start and end tokens of entity pairs for each relation category. \cite{shang2022onerel} reformulates joint extraction as fine-grained triple classification, enabling the task to simultaneously capture information pertaining to head entities, relations, and tail entities. \cite{tang2022unirel} propose an interaction graph constructed on top of standard self-attention mechanisms to unify interaction modeling. The table filling method is constrained by a fixed table size and may struggle to accommodate relational triple extraction in lengthy sentences.

\section{Methodology}
\label{sec:methodology}

This section begins by presenting the task definition in Section \ref{sec:task definition}. Afterwards, we describe our model's implementation in Section \ref{sec:interaction network}.

\subsection{Task Definition}
\label{sec:task definition}

Consider a sentence $S = \{w_1, w_2, \ldots, w_N\}$ with length $N$. The objective of relational triple extraction is to identify all possible relational triples $\mathcal{T} = \{(h_i, r_i, t_i)\}^{\mu}_{i=1}$ contained in $S$, where $\mu$ represents the total number of triples. Here, $h_i, t_i \in E$, with $E$ denoting the entity set within $S$. Meanwhile, $r_i \in R$, and $R = \{r_1, \ldots, r_\mathcal{K}\}$ denotes the set of $\mathcal{K}$ predefined relation categories.

Let $\mathcal{T}_{tra}$ be the relational triple set in a training dataset. For a triple $(h, r, t)$ from the testing dataset, the three categories, \emph{entirely seen}, \emph{partially seen} and \emph{unseen}, are specified below:

\begin{itemize}
	\item The triple $(h, r, t)$ is categorized as \emph{entirely seen} type, when $(h, r, t) \in \mathcal{T}_{tra}$.
	\item The triple $(h, r, t)$ fall into the \emph{partially seen} type, if it satisfies either $(h, r, \cdot) \in \mathcal{T}_{tra}$ or $(\cdot, r, t) \in \mathcal{T}_{tra}$. 
	\item Otherwise, if $(h, r, t) \notin \mathcal{T}_{tra}$ and $(h, r, \cdot) \notin \mathcal{T}_{tra}$ and $(\cdot, r, t) \notin \mathcal{T}_{tra}$, the triple is categorized as \emph{unseen}.
\end{itemize}

\subsection{Interaction Network}
\label{sec:interaction network}

\begin{figure}
	\centering
	\includegraphics[width=0.85\linewidth]{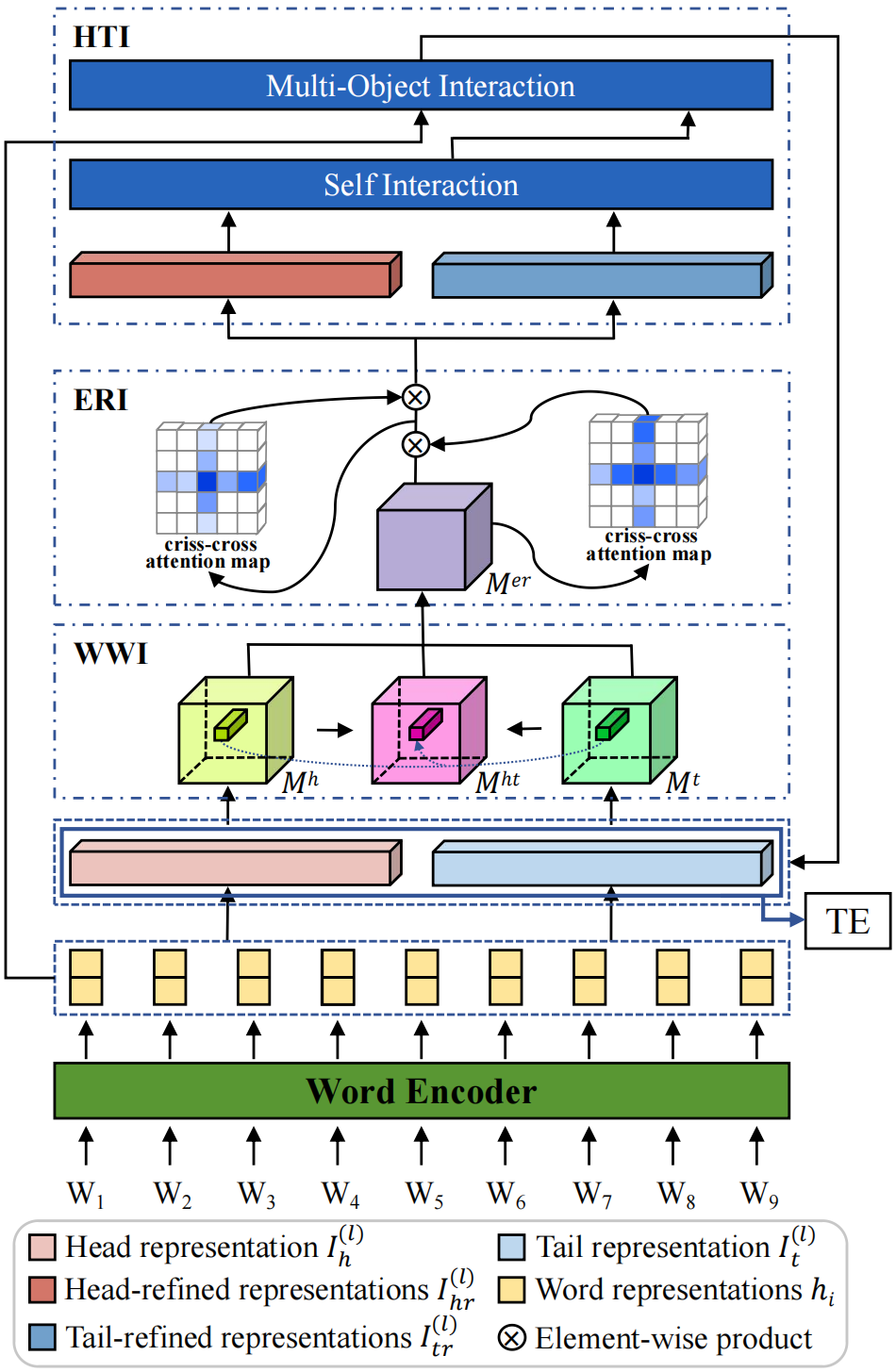}
	\caption{Model Architecture.}
	\label{fig:model}
\end{figure}

Figure \ref{fig:model} illustrates the architecture of our model designed for relational triple extraction.

The proposed model consists of five primary components: an \emph{Encoder} module, a \emph{Word-Word Interaction} (WWI) module, an \emph{Entity-Relation Interaction} (ERI) module, a \emph{Head-Tail Interaction} (HTI) module, and a \emph{Triple Extraction (TE)} module. The interaction network is composed by the three WWI, ERI, and HTI modules.

Given an input, it is processed by the interaction network multiple times in an iterative manner to progressively refine the representations corresponding to head and tail entities. Finally, the \emph{TE} module builds the \emph{Interaction Map} using these refined representations to extract all triples from a sentence.

Comparing with existing models, our interaction network for extracting relational triples, aims to fully leverage all potential semantic interactions relevant to the parameters of a relational triple. It enables the extraction of entirely seen, partially seen, and unseen relational triples. Unlike existing table-filling methods \citep{wang2020tplinker, ren2021novel, shang2022onerel, dai2024soirp} that rely solely on head and tail entity semantics, our model integrates different levels of semantic interaction (WWI, ERI, and HTI). Traditional methods usually place excessive reliance on the model's memory capabilities while overlooking its generalization abilities, making them inefficient in real-world applications. Our approach encompasses all types of relational triples, including entirely seen, partially seen, and unseen relational triples, making it more suitable for practical applications.

\subsubsection{Encoder}
\label{sec:encode}

For an input sentence $S=\{w_1, w_2, \ldots, w_N\}$ along with a predefined relation set $R = \{r_1, \ldots, r_\mathcal{K}\}$, to lean semantic features of tokens and build the semantic dependencies between them, a pre-trained language model (BERT \citep{eberts2020span}) is utilized to convert the sentence into a sequential abstract representation. The process is formalized as follows:

\begin{equation}
	\begin{aligned}
		[h_1, \ldots, h_{N + \mathcal{K}}] & = \text{BERT}\left([w_1, \ldots , w_N, r_1, \ldots , r_\mathcal{K}]\right) \\
		\text{or},\ \	H & = \text{BERT}\left( S \oplus R\right)
	\end{aligned}
\end{equation}
where, $\oplus$ denotes the string concatenation operation, which indicates that all relation type labels are concatenated with the sentence. It is effective to capture the semantic interplay between tokens and relation types. The output $H= [h_1, \ldots, h_{N + \mathcal{K}}] \in \mathbb{R}^{(N+\mathcal{K}) \times d_e}$ is the abstract sentence representation.

Then, $H$ is fed into two independent \emph{Multilayer Perceptrons} (MLP) to generate the initial head and tail representations, denoted as $I_h ^{(1)}$ and $I_t^{(1)}$, respectively. It is shown in the following equations:
\begin{equation}
	\begin{aligned}
		I_h ^{(1)} & = W_hH + b_h \\
		I_t ^{(1)} & = W_tH + b_t \\
	\end{aligned}
\end{equation}

\noindent where $W_{h/t} \in \mathbb{R}^{d_e \times d_e}$ and $b_{h/t} \in \mathbb{R}^{d_e}$ are trainable weights and biases, respectively.

\subsubsection{Word-Word Interaction}

The Word-Word Interaction plays a vital role in capturing the contextual relationships between words within a sentence. It is essential for capturing the overall semantics. This interaction reinforces semantic features of elements in a two-dimensional sentence representation, which makes the full use of relation and entity information. By analyzing these word relationships, the model can discern subtle nuances in meaning, allowing it to identify relational triples more accurately. This capability is particularly beneficial when dealing with relational triples unseen during training.

To better learn the word-word interaction information, the one-dimensional head and tail representations are transformed into two two-dimensional head and tail representations, denoted $M^h \in \mathbb{R}^{(N + \mathcal{K}) \times (N + \mathcal{K}) \times d_e}$ and $M^t \in \mathbb{R}^{(N + \mathcal{K}) \times (N + \mathcal{K}) \times d_e}$, respectively. The two-dimensional representations are similar to the existing table filling method \citep{wang2020tplinker, shang2022onerel, ning2023od}. The cell $(w_i, w_j)$ in each two-dimensional representation, $M^{\epsilon}_{i, j}$ $ \epsilon\in \{h, t\}$ is calculated as follows:
\begin{equation}
	M^{\epsilon}_{i, j} = ReLU(W_{c}[h_i;h_j] + b_{c}) \\
\end{equation}

\noindent with $1 \leq i, j \leq N$, where $ReLU(\cdot)$ denotes the ReLU nonlinearity \citep{agarap2018deep}, and ``;'' represents the concatenation operation. $W_c \in \mathbb{R}^{d_e \times 2d_e}$ and $b_c \in \mathbb{R}^{d_e}$ are trainable parameters.

The dimensionality expansion process of the head and tail representations involves word-to-word interactions. To further enhance word-word interactions, we employ a tensor-based transformation that derives the head-tail semantic representation.

For two cells $z_i, z_j \in \mathbb{R}^{d_e}$ in two-dimensional head and tail representations $M^{h/t}$, the tensor-oriented operation is formulated as:
\begin{equation}
	\mathcal{M}\left(z_i, z_j ; V\right)=z_i^{\top} V^{[1: m]} z_j
\end{equation}

\noindent where $V \in \mathbb{R}^{d_e \times d_e \times m}$ is a tensor. For each slice $V^{[k]} \in \mathbb{R}^{d_e \times d_e}$, this operation computes the dot product of two cell vectors in a given vector space:

\begin{equation}
	\mathcal{M}\left(z_i, z_j ; V\right)_k=z_i^{\top} V^{[k]} z_j
\end{equation}

Through the use of multiple vector spaces, the proposed tensor operation is capable of thoroughly capturing token-token interactions.

Finally, we construct a two-dimensional entity-relation representation through a nonlinear projection:
\begin{equation}
	\begin{aligned}
		M^{er} & = f(\text{Linear}([M^h;M^{ht};M^t]))  \\
		M^{ht} & = \mathcal{M}\left(z_i, z_j ; V\right) \in \mathbb{R}^{N \times N \times m}  \label{equ:m}\\
	\end{aligned}
\end{equation}

\noindent here, $f(\cdot)$ represents the GELU nonlinearity \citep{hendrycks2016gaussian}. The dimension of $M^{er}$ is denoted as $d_r$, $M^{er} \in \mathbb{R}^{N \times N \times d_r}$.

\subsubsection{Entity-Relation Interaction}
\label{Entity-Relation Interaction}

The entity-relation interaction centers on how entities connect with various types of relations, which is crucial for contextualizing them within their specific relational frameworks. This interaction captures long-distance semantic dependencies between relations and entities, allowing the model to filter out the semantic features related to redundant relational triples. By emphasizing these interactions, the model can effectively differentiate between similar entities that may have distinct relationships in different contexts, thus enhancing the precision of the extracted triples.

\begin{figure}
	\centering
	\includegraphics[width=0.85\linewidth]{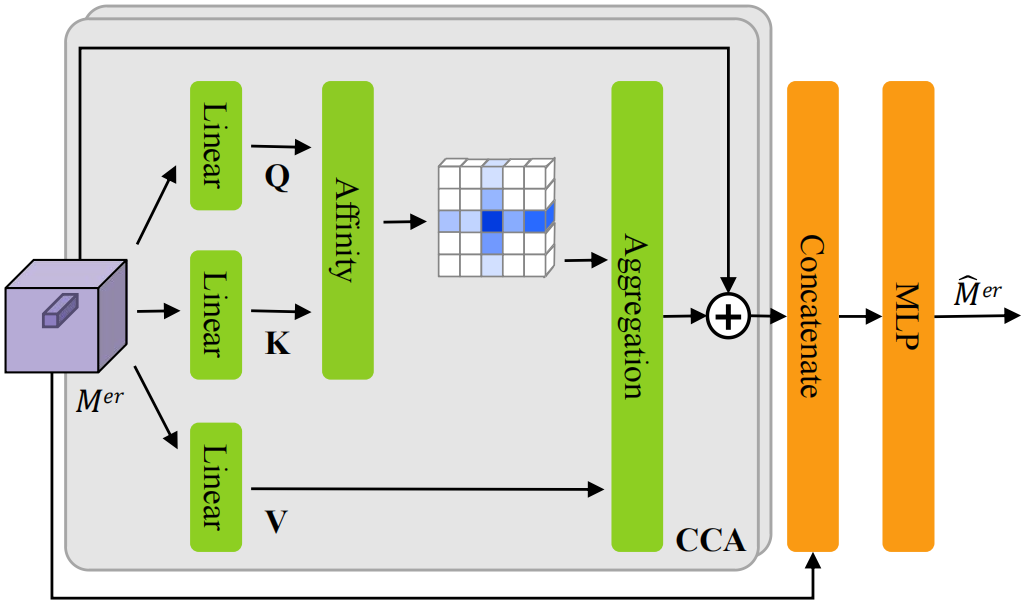}
	\caption{The details of Entity-Relation Interaction.}
	\label{fig:cca}
\end{figure}

There are potential dependencies between elements of an entity-relation representation $M^{er}$. For example, if an element in the Entity-Entity region is labeled, it must be labeled correctly in both the Head-Relation and the Relation-Tail region. Notably, we observed that the positions of the Entity-Entity, Head-Relation, and Relation-Tail markers exhibit a criss-crossing long-distance semantic dependency. To model these dependencies, we utilize the \emph{Criss-Cross Attention} (CCA) \citep{huang2019ccnet} to encode this entity-relation representation matrix, as shown in Figure \ref{fig:cca}.

Given an entity-relation representation $M^{er} \in \mathbb{R}^{N \times N \times d_r}$, unlike the \emph{Criss-Cross Attention} used by \cite{huang2019ccnet}, two fully connection layers are first applied on $M^{er}$ to produce a pair of feature maps $Q$ and $K$ with $\{Q, K\} \in \mathbb{R}^{N \times N \times d_r}$. The motivation for using $1 \times 1$ filters instead of the traditional convolutional layer stems from the fact that the fully connected layer has strong nonlinear fitting capabilities for learning task-relevant semantic features.

After computing $Q$ and $K$, the $\mathbf{Affinity}$ operation yields an attention map $A \in \mathbb{R}^{(N + N - 1) \times (N \times N)}$.  Each $u$ (spatial position in $Q$) gives a vector $Q_u \in \mathbb{R}^{d_r}$. Simultaneously, a set $\Omega_u \in \mathbb{R}^{(N + N - 1) \times d_r} $ is derived via selecting feature vectors from $K$ that share the same row or column with position $u$. Here, $\Omega_{i,u} \in \mathbb{R}^{d_r}$ denotes the $i$-th component of $\Omega_u$. The $\mathbf{Affinity}$ operation is formulated as:
\begin{equation}
	d_{i, u} = Q_{u} \Omega_{i, u}^{\top} / \sqrt{d_r}, 
\end{equation}

\noindent where $d_{i,u}$ represents the correlation score between $Q_u$ and $\Omega_{i,u}$. $i$ ranges from 1 to $N+N-1$, and $D \in \mathbb{R}^{(N + N - 1) \times (N \times N)}$.

Then, a softmax function is performed over $D$ along the channel axis to obtain the attention map $A$.

An additional fully connected layer is applied to $M^{er}$ to obtain $V \in \mathbb{R}^{N \times N \times d_r}$ for feature adaptation. At every spatial position $u$ within $V$, we extract a vector $V_u \in \mathbb{R}^{d_r}$ along with a set $\Phi_u \in \mathbb{R}^{(N + N - 1) \times d_r}$. $\Phi_u$ consists of feature vectors from $V$ lying on the same row or column with position $u$. The surrounding information is then integrated via an $\mathbf{Aggregation}$ operation, which is defined as:
\begin{equation}
	\hat{M}^{er}_{u}=\sum_{i=0}^{N+N-1} A_{i, u} \Phi_{i, u} + M^{er}_{u}
\end{equation}

\noindent where $\hat{M}^{er}_{u}$ is the feature vector from $\hat{M}^{er} \in \mathbb{R}^{N \times N \times d_r}$ located at position $u$, and $A_{i,u}$ is the scalar entry in $A$ for channel $i$ at the same position. This allows the contextual information to be fused with the local feature $M^{er}$, thereby strengthening the semantic dependencies within the entity-relation representation. To fully utilize the potential of CCA, CCA unrolled into 2 loops. For more details, please referred to \cite{huang2019ccnet}.

Finally, we use  concatenation operators and an MLP to produce an enhanced entity-relation representation, which is defined in the following equation:
\begin{equation}
	\begin{aligned}
		\hat{M}^{er} & = \text{CCA}(M^{er})_{\times 1/2} \\
		\hat{M}^{er} & = W_{er}[\hat{M}^{er};M^{er}] + b_{er} \\
	\end{aligned}
\end{equation}

\noindent where, ``$\times 1/2$'' means CCA unrolled into 1 or 2 loops, $[\cdot;\cdot]$ is concatenation operators, $W_{er} \in \mathbb{R}^{d_r \times 2d_r}$ and $b_{er} \in \mathbb{R}^{d_r}$ are trainable parameters.

\subsubsection{Head-Tail Interaction}
\label{Head-Tail Interaction}

The head-tail interaction is crucial for comprehending how relational triples are structurally composed, since it involves interpreting both head and tail entities in light of their relationship. This interaction enhances the global semantic features associated with relational triples, aiding in the remediation of triples that may have been overlooked or omitted in earlier steps. Consequently, it ensures that the model can accurately associate the correct entities with their respective relations, especially in complex sentences where relationships may not adhere to straightforward patterns.

The \emph{Head-Tail Interaction Module} is proposed to address the remediation of overlooked or omitted triples. This module comprises two key components: \emph{Self Interaction} and \emph{Multi-Object Interaction}. The \emph{Self Interaction} component can utilize either MLP or \emph{Multi-Head Self-Attention (MHA)}, depending on the specific task requirements. Meanwhile, the \emph{Multi-Object Interaction} component can employ either MLP or \emph{Cross Attention (CA)}. Specifically, MHA is a mechanism that allows the model to extract multiple relationships from the input through parallel attention processes, while CA is designed to align and relate information from diverse sources or modalities. Finally, MLP is a robust technique for modeling complex, non-linear interactions by leveraging fully connected layers.

Assuming that the current iteration is $l$, after WWI and ERI, the entity-relation representation contains rich information about the interactions linking entity pairs with their corresponding relations.
In this module, we first use a Maxpooling operation and an MLP for $\hat{M}^{er}$ to generate the head-refined representation $I_{hr}^{(l)}$ and tail-refined representation $I_{tr}^{(l)}$ respectively, as shown in Equation \ref{eq:head and tail representation}.
\begin{equation}
	\begin{aligned}
		\label{eq:head and tail representation}
		I_{hr}^{(l)} & = W_{hr}\text{Maxpool}_{hr}(\hat{M}^{er}) + b_{hr}  \\
		I_{tr}^{(l)} & = W_{tr}\text{Maxpool}_{tr}(\hat{M}^{er}) + b_{tr}  \\
	\end{aligned}
\end{equation}

\noindent where $W_{hr/tr} \in \mathbb{R}^{d_r \times d_e}$ denotes learnable weight matrices, and $b_{hr/tr} \in \mathbb{R}^{d_e}$ represents learnable bias vectors.

Then, we employ the \emph{Self Interaction} on $I_{h/t}^{(l)}$ to extract interaction cues from the head-relation and tail-relation pairs. The formula is as follows:
\begin{equation}
	\tilde{I}_{hr/tr}^{(l)} =\text{Self-Interaction}_{\varepsilon} \left(I_{hr/tr}^{(l)}\right) \\
\end{equation}
\noindent where $\varepsilon \in \{\text{MHA}, \text{MLP}\}$ denotes the two strategies of the \emph{Self Interaction}. Each strategy has its advantages depending on the dataset and task. However, MHA tends to perform poorly on long-sentence, multi-relation datasets, as shown in Section \ref{Influence of the Head-Tail Interaction}. On the other hand, MLP, while simpler, can offer more stable performance on these datasets by processing information through fully connected layers, which helps to improve the performance on complex sentences.

Next, to prevent the omission of certain global and local semantic details, we adopt the \emph{Multi-Object Interaction} approach to allow head-refined representation and tail-refined representation to interact with $H$, so as to extract the information about the reasoning paths between relational triples. The computation procedure is given below:

\begin{equation}
	\begin{aligned}
		\tilde{I}_{hr/tr}^{(l+1)} & = \text{MulObj-Interaction}_{\delta}\left(\tilde{I}_{hr/tr}^{(l)}, H\right) \\ & = \begin{cases}\text{CrossAtt}\left(\tilde{I}_{hr/tr}^{(l)}, H, H\right) & \delta \in \text{CA} \\ 
			\text{MLP}\left(\tilde{I}_{hr/tr}^{(l)}; H\right) & \delta \in \text{MLP}\end{cases}
	\end{aligned}
\end{equation}
\noindent where $\delta \in \{\text{CA}, \text{MLP}\}$ represents the two strategies for the Multi-Object Interaction process. In the first case, CA allows for interactions between the head and tail representations through attention mechanisms, leveraging the information from $H$ for deeper reasoning. In the second case, MLP applies a simpler fully connected layer to combine the refined representations with $H$, offering a more straightforward interaction mechanism. Both approaches aim to improve the extraction of reasoning paths across relational triples, but their effectiveness depends on the specific dataset.

Finally, we generate new head representation $I_{h}^{(l+1)}$ and tail representation $I_{t}^{(l+1)}$ with an MLP model. These representations are fed back to WWI for the subsequent iteration. It is worth noting that the network parameters are reused across all iterations.

In summary, the new head and tail representations are formulated via the equation below \ref{eq:HTI}.
\begin{equation}
	\begin{aligned}
		\label{eq:HTI}
		I_{h/t}^{(l+1)} & =\operatorname{ReLU}\left(\tilde{I}_{hr/tr}^{(l+1)} W+b\right) \\
		I_{h/t}^{(l+1)} & =\operatorname{LN}\left(I_{h/t}^{(l+1)} + I_{h/t}^{(l)}\right)
	\end{aligned}
\end{equation}

\noindent where $\operatorname{LN}(\cdot)$ is Layer Norm \citep{ba2016layer}. A residual connection avoids gradient vanishing.

\subsubsection{Triple Extraction}

In this module, an \emph{Interaction Map} is generated through the interaction of head and tail representations. Assuming that the number of final iterations of the interaction network is $L$, the head representation $I_h^{(L)}$ and the tail representation $I_t^{(L)}$ are fed into the Biaffine \citep{yu2020named} for generating the \emph{Interaction Map}. The equation is given below:
\begin{equation}
	\mathcal{I}_{ij} = I_h^{(L){\top}} U_B I_t^{(L)} + W_B(I_h^{(L)};I_t^{(L)}) + b_B \\
\end{equation}

\noindent where, $\mathcal{I} \in \mathbb{R}^{(N + \mathcal{K}) \times (N + \mathcal{K})}$. $U_B$ is a $d_e \times 1 \times d_e$ tensor, $W_B$ is a $2d_e \times 1$ matrix and $b_B$ is the bias.

The \emph{Interaction Map} strategy \citep{tang2022unirel} stems from its narrowed prediction space, estimated at $O((N + \mathcal{K})^2)$, which is notably smaller compared to other methods requiring $O(N \times \mathcal{K} \times N)$ complexity for triple prediction. In addition, it also has an efficient decoding efficiency. Although OD-RTE \citep{ning2023od} enhances the tagging scheme based on OneRel \citep{shang2022onerel} and introduces the Bi-DW decoding algorithm, its prediction space remains at $O(N \times \mathcal{K} \times N)$.

\begin{figure}
	\centering
	\includegraphics[width=0.85\linewidth]{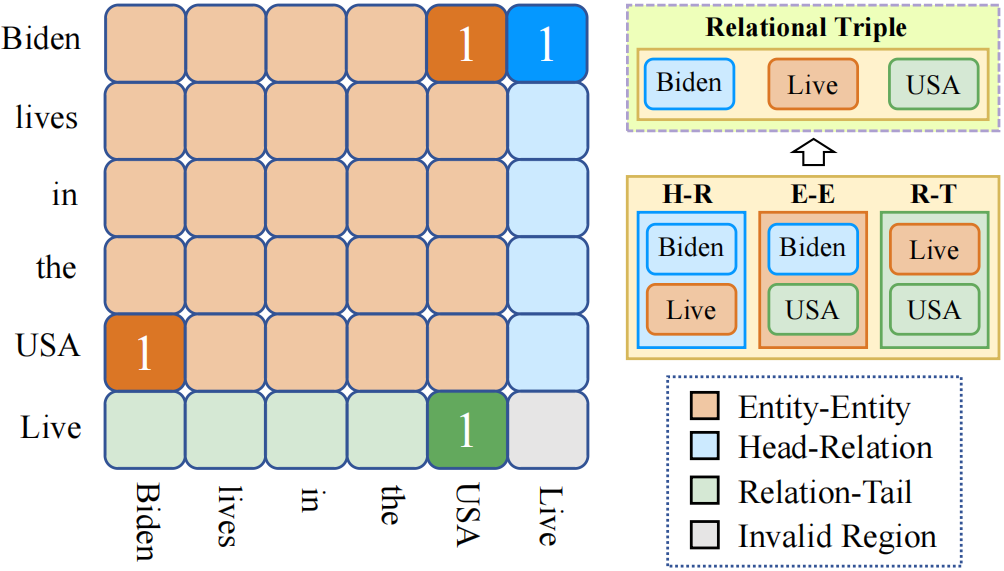}
	\caption{An Example of entity-pair tagging.}
	\label{fig:IM}
\end{figure}

The \emph{Interaction Map} fuses Entity-Entity (E-E) table, Head-Relation (H-R) table, Relation-Tail (R-T) table in the same table for labeling. Consider the example ``Biden lives in the USA.'', which is visualized in Figure \ref{fig:IM}.
 
The triple (Biden, live\_in, USA) can be determined if (Biden, USA) is tagged in the Entity-Entity section, (Biden, live\_in) is tagged in the Head-Relation section, and (live\_in, USA) is tagged in the Relation-Tail section. Thus, the \emph{Interaction Map} is of size $(N + \mathcal{K}) \times (N + \mathcal{K})$ (see \cite{tang2022unirel} for details).

In decoding, we first obtain all valid head/tail entities per relation from the Head-Relation and Relation-Tail regions. Then, for each relation, we examine candidate entity pairs after pruning by the Entity-Entity region.

The loss function for our framework is formulated as:
\begin{equation}
	\begin{array}{r}
		\mathcal{L}=-\frac{1}{(N+\mathcal{K})^2} \sum_i^{N+\mathcal{K}} \sum_j^{N+\mathcal{K}}\left(\mathcal{I}_{ij}^* \log \mathcal{I}_{ij}\right. \\
		\left.+\left(1-\mathcal{I}_{ij}^*\right) \log \left(1-\mathcal{I}_{ij}\right)\right)
	\end{array}
\end{equation}

\noindent where $\mathcal{I}^*$ denotes the ground-truth Interaction Map matrix.

\section{Experiment}
\label{sec:experiment}

\subsection{Experimental Settings}
\label{Experimental Settings}

\begin{table*}[h]
	\centering
	\caption{\label{table:datasets} Statistics of the datasets.}
	\begin{tabular}{lcccc c cccc}
		\hline
		\multirow{2}{*}{ Category } & \multicolumn{4}{c}{ Details of the dataset } & & \multicolumn{4}{c}{ Type of triple }\\
		&  Train  &  Valid  &  Test  &  Relation & &  Entire  &  Partial  &  Unseen  &  Others \\
		\hline 
		$\text{NYT}$-$\text{R}$ & 56196 & 5000 & 5000 & 24 & & 348 & 3307  & 886  & 459  \\
		$\text{WebNLG}$-$\text{R}$ & 5019 & 500 & 703 & 216 & & 155  & 178  & 174  & 196\\
		$\text{NYT}$ & 56195 &  5000  & 5000 & 24 & & 4292  & 473  & 88  & 147  \\
		$\text{WebNLG}$ & 5019 & 500  & 703 & 171 & & 580  & 42  & 17  & 64  \\
		\hline
	\end{tabular}
\end{table*}

\begin{table*}[h]
	\centering
	\caption{\label{table:overlapping of datasets} Overlapping patterns in datasets. $N$ denotes the total count of relational triples.}
	\begin{tabular}{lccccccccc}
		\hline
		\multirow{2}{*}{ Dataset } & \multicolumn{9}{c}{ Overlapping Pattern }  \\
		\cline { 2 - 10 }
		& Normal & SEO & EPO & SOO  & $N=1$ & $N=2$ & $N=3$ & $N=4$ & $N \geq 5$ \\
		\hline
		$\text{NYT}$-$\text{R}$ & 3965 & 709 & 408 & 110   & 3958& 733& 192& 50& 48 \\
		$\text{WebNLG}$-$\text{R}$ & 389 & 293 & 9 & 84   & 411& 143& 81& 58& 10 \\
		$\text{NYT}$ & 3266 & 1297 & 978 & 45   & 3,244& 1,045& 312& 291& 108 \\
		$\text{WebNLG}$ & 245 & 457 & 26 & 84    & 266& 171& 131& 90& 45 \\
		\hline
	\end{tabular}
\end{table*}

\textbf{Datasets}: we assess our method using two benchmark datasets introduced by \citep{lee2022does}: NYT-R and WebNLG-R. These datasets, derived from the NYT dataset \citep{riedel2010modeling} and the WebNLG dataset \citep{gardent2017creating}, are primarily designed to assess relational triple extraction performance on ``unseen'' entities. The NYT-R dataset was constructed colle by aligning relations in Freebase articles in the New York Times corpus. It contains 24 predefined relations. On the other hand, the WebNLG-R dataset is initially devised for the Natural Language Generation Challenge. It is annotated with 171 predefined relations types.

Table \ref{table:datasets} provides the statistical breakdown of these datasets, which are categorized into four groups: Entire, Partial, Unseen, and Others. The Entire, Partial and Unseen indicate that all relation triples in the sentence are of the corresponding relational triple type. The Others suggest that relational triples may contain Entire, Partial or Unseen types of relational triples. According to \citep{zheng2021prgc}, SEO stands for Single Entity Overlapping, EPO refers to Entity Pair Overlapping, and SOO denotes Subject Object Overlapping. To maintain consistency with prior research \citep{zhan2022simple, sui2023joint, ning2023od}, we adopt the conventional micro Precision (Prec.), Recall (Rec.), and F1-score metrics for gauge model performance.

\textbf{Baseline}: For comparison, we used the following model as a baseline: (1)$\mathrm{CasRel}$ \citep{wei2020novel} first identifies all possible head-entities within a sentence. Subsequently, for each head-entity, it determines all potential relations and their corresponding tail-entities. (2)$\mathrm{TPlinker}$ \citep{wang2020tplinker} can identify overlapping relations that share either one or two entities, and it remains unaffected by exposure bias. (3)$\mathrm{PRGC}$ \citep{zheng2021prgc} approaches the task from a new angle by breaking it down into three subtasks: Relation Judgement, Entity Extraction, and Subject-object Alignment. Subsequently, it introduces a joint extraction framework based on Potential Relation and Global Correspondence. (4)$\mathrm{BiRTE}$ \citep{ren2022simple} operates as a bidirectional extraction framework. It generates triples by leveraging entity pairs obtained from two complementary directions. (5)$\mathrm{OneRel}$ \citep{shang2022onerel} reinterprets joint extraction as a fine-grained triple classification task. Its architecture comprises a scoring-based classifier along with a relation-specific horns tagging strategy. (6)$\mathrm{SPN}$ \citep{sui2023joint} frames joint entity and relation extraction as a direct set prediction task. This allows the extraction model to avoid the difficulty of determining the order of multiple triples. (7)$\mathrm{UniRel}$ \citep{tang2022unirel} unifies the representations of entities and relations and unify the modeling of interactions with a Interaction Map. (8)$\mathrm{RTF}$ \citep{an2024rtf} introduces a novel region-based tagging scheme along with a bi-directional decoding strategy. This approach treats every relational triple as a region within a relation-specific table, and extracts triples by locating the two endpoints of each region. (9)$\mathrm{SOIRP}$ \citep{dai2024soirp} comprises a Subject-Object Interaction module, which is built from multiple distinct interactions, along with a Reasoning Path Extraction module inspired by the Transformer decoder architecture. 

\textbf{Implementation Details}: for fair comparison purposes, the Bert-base-cased language model \footnote{Available at https://huggingface.co/bert-base-cased} is used. It features 12 Transformer blocks and a hidden size of $d = 768$. All parameters are optimized via the Adam optimizer \citep{kingma2014adam} on two rearranged datasets, employing a learning rate of $\text{3e-5}$ and batch sizes of 6/4. In equation \ref{equ:m}, the hidden size of $d_r = 512$ and the vector space based on tensor operations is $m=5$. In Section \ref{Entity-Relation Interaction}, CCA unrolled into 1/2 loops on $\mathrm{NYT}$-$\mathrm{R}$/$\mathrm{WebNLG}$-$\mathrm{R}$ rearrange datasets. In Section \ref{Head-Tail Interaction}, Self Interaction employed MHA/MLP and Multi-Object Interaction adopt CA/MLP on $\mathrm{NYT}$-$\mathrm{R}$/$\mathrm{WebNLG}$-$\mathrm{R}$ rearrange datasets. Additionally, the interaction network is iterated only once, and sentences are truncated to a maximum length of 100, in accordance with previous work. (Note that this limit excludes the number of relations.). All experiments ran on a single NVIDIA RTX A6000 GPU.

\subsection{Main Results}

\begin{table*}[h]
    \small
    \centering
	\caption{\label{table:main result} Main results.}
    \resizebox{\linewidth}{!}{
        \begin{tabular}{l cccccc c cccccc}
            \hline
            \multirow{2}{*}{ Model } & \multicolumn{6}{c}{ $\mathrm{NYT}$-$\mathrm{R}$ }  &    & \multicolumn{6}{c}{ $\mathrm{WebNLG}$-$\mathrm{R}$ } \\
            & Prec. & Rec. & F1 & Entire & Partial & Unseen &   & Prec. & Rec. & F1 & Entire & Partial & Unseen \\
            \hline
            $\mathrm{CasRel}^\dagger$ \citep{wei2020novel}  & 65.9 & 60.1 & 62.9 & 85.8 & 65.0 & 42.3   &   & 73.6 & 64.2 & 68.6 & 89.6 & 52.3 & 41.5 \\
            $\mathrm{TPlinker}^\dagger$ \citep{wang2020tplinker}  & 69.0 & 60.8 & 64.7 & 83.3 & 66.7 & 46.8   &   & 75.1 & 63.9 & 69.1 & 88.5 & 52.7 & 42.9  \\
            $\mathrm{PRGC}^\dagger$ \citep{zheng2021prgc}  & 63.5 & 61.6 & 62.6 & 81.6 & 64.2 & 45.1   &   & 61.6 & 62.0 & 61.8 & 79.2 & 47.2 & 28.3  \\
            $\mathrm{BiRTE}^\S$ \citep{ren2022simple} & 64.5 & 65.4 & 64.9 & 86.1 & 67.0 & 45.2 &     & 71.0 & 64.6 & 67.6 & 89.2 & 53.9 & 36.7  \\
            $\mathrm{OneRel}^\S$ \citep{shang2022onerel}  & $\underline{70.8}$ & 58.2 & 63.9 & 82.5 & 66.5 & 44.2   &   & 75.9 & 66.5 & 70.9 & $\underline{90.8}$ & $\underline{57.2}$ & 39.2  \\
            $\mathrm{SPN}^\S$ \citep{sui2023joint} & 68.4 & 61.2 & 64.6 & 84.8 & 68.2 & 48.5 &     & 64.1 & 60.2 & 62.1 & 81.3 & 53.3 & \underline{44.4}  \\
            $\mathrm{UniRel}^\S$ \citep{tang2022unirel} & \textbf{74.2} & 57.1 & 64.6 & 81.2 & 66.8 & 44.8 &     & 72.9 & 64.0 & 68.2 & 85.7 & 51.3 & 43.3  \\
            $\mathrm{RTF}$ \citep{an2024rtf}  & 68.8 & 62.3 & 65.4 & - & - & - &     & \underline{76.5} & $\underline{66.5}$ & \underline{71.2} & - & - & -  \\
            $\mathrm{SOIRP}^\S$ \citep{dai2024soirp} & 68.5 & \underline{66.5} & \underline{67.5} & \underline{86.9} & \underline{69.8} & \underline{49.5} &     & 75.2 & 62.9 & 68.5 & 90.5 & 53.9 & 38.2  \\
            \hline
            $Ours$  & 70.2 & $\mathbf{67.1}$ & $\mathbf{68.6}$ & $\mathbf{89.1}$ & $\mathbf{70.8}$ & $\mathbf{52.6}$  &  & $\mathbf{78.1}$ & $\mathbf{67.8}$ & $\mathbf{72.6}$ & $\mathbf{90.8}$ & $\mathbf{57.2}$ & $\mathbf{58.7}$ \\
            \hline
        \end{tabular}
    }
\end{table*}

We evaluate our model against nine competitive baselines. Table \ref{table:main result} presents the experimental outcomes on two rearranged datasets, where the top performance is highlighted in bold and the runner-up is underlined. The performance of Entire, Partial and Unseen metrics is given in F1 scores. \S \ means that the results are not reported in the paper, while $\dagger$ denotes that the results are reported in \cite{lee2022does}. They are obtained through re-implemented its source code with the same evaluated datasets. The experimental findings reveal that:

(1) Some models have higher F1 scores, but it does not balance the performance between seen and unseen relational triples. For instance, the $\mathrm{OneRel}$ model achieves an F1 score 1.8\% points higher than the $\mathrm{TPlinker}$ model on the $\mathrm{WebNLG}$-$\mathrm{R}$ dataset, but it is 3.7\% points lower in the Unseen entity triples. OneRel's effectiveness is largely tied to the availability of annotated entities in the training data. Its performance easily overfitting to the seen relational triples. Therefore, the final performance is strongly influenced by the types of relational triples encountered during training.

(2) Our model surpasses all compared baseline methods and achieves the highest F1 results on the two rearranged datasets. This underscores our approach's efficacy in handling entirely seen, partially seen and unseen relational triples, a capability attributed to its effective interaction network. This network not only utilizes word-word, entity-relation and head-tail interactions in the relational triples, but also integrates the information on these effective interactions. As a result, the interaction network encodes rich entity and relation semantic features. They are effective to improve the performance.

(3) The experimental results show that our model surpasses all compared baseline models, achieving the best performance across the Entire, Partial, and Unseen metrics on the rearranged dataset. In the Entire metric, the model may encounter relational triples during training, demonstrating its capacity for memorization. In the Partial and Unseen metrics, the model encountered relational triples that were partially seen or unseen during training, highlighting its strong generalization capabilities. This performance is attributed to the integration of word-word, entity-relation and head-tail interaction information, which gives it the ability to memorize and generalize.

\subsection{Ablation Study}

\begin{table*}[h]
    \centering
    \small
	\caption{\label{table:ablation Study} Ablation study.}
    \begin{tabular}{l cccccc c cccccc}
        \hline
        \multirow{2}{*}{ Model } & \multicolumn{6}{c}{ $\mathrm{NYT}$-$\mathrm{R}$ }  &    & \multicolumn{6}{c}{ $\mathrm{WebNLG}$-$\mathrm{R}$ } \\
        & Prec. & Rec. & F1 & Entire & Partial & Unseen &   & Prec. & Rec. & F1 & Entire & Partial & Unseen \\
        \hline
        Fully Model  & 70.2 & $\mathbf{67.1}$ & $\mathbf{68.6}$ & \underline{89.1} & $\mathbf{70.8}$ & $\mathbf{52.6}$  &  & $\mathbf{78.1}$ & $\mathbf{67.8}$ & $\mathbf{72.6}$ & $\mathbf{90.8}$ & $\mathbf{57.2}$ & $\mathbf{58.7}$ \\
        \quad ${w/o\ \mathrm{WWI}}$ & 70.3 & \underline{65.0} & \underline{67.6} & 88.4 & \underline{69.5} & 50.7     &     & 77.1 & 63.5 & 69.7 & 87.0 & \underline{55.9} & 50.0 \\
        \quad ${w/o\ \mathrm{ERI}}$ & 69.4 & 65.0 & 67.1 & $\mathbf{89.3}$ & 69.2 & $\underline{51.7}$     &    & 76.8 & \underline{67.0} & 71.6 &  90.0 & 54.7 & 55.8 \\
        \quad ${w/o\ \mathrm{HTI}}$ & 70.6 & 64.0 & 67.2 & 87.9 & 69.4 & 50.4     &    & 75.8 & $\mathbf{67.8}$ & 71.6 & \underline{90.5} & 55.9 & \underline{56.4} \\
        \quad ${w/o\ \mathrm{ALL}}$ & $\underline{74.2}$ & 57.1 & 64.6 & 81.2 & 66.8 & 44.8     &    & 68.1 & 52.8 & 59.5 & 75.7 & 43.5 & 43.5 \\
        \quad ${w/o\ \mathrm{SEP}}$  & $\mathbf{74.8}$ & 54.2 & 62.8 & 76.0 & 65.9 & 45.2 &     & $\underline{77.7}$ & 66.5 & \underline{71.7} & 90.4 & 53.4 & 52.1  \\
        \hline
    \end{tabular}
\end{table*}

In this section, we conducted ablation experiments to assess the impact of each component of the model. The highest scores are in bold and second scores are in underline. The performance of Entire, Partial and Unseen metrics is given in F1 scores.

(1) \emph{w/o WWI} denotes the model where word-word interaction is removed from the interaction network. In this setting, the model cannot make full use of relation and entity information. Experimental results reveal a slight performance degeneration on the $\mathrm{NYT}$-$\mathrm{R}$ dataset, with a reduction of 1.0\% in the F1-score. The degeneration suggests that word-word interactions are vital for capturing contextual relationships between words, thereby enhancing the discriminability of the model. The performance on the $\mathrm{WebNLG}$-$\mathrm{R}$ dataset shows a more significant decline, with a 2.9\% decrease in F1 score. This highlights the critical role of word-word interactions in reinforcing the semantic features of relational triples.

By comparison, the performance of  \emph{w/o WWI} on the Unseen metric is notably lower than the full model, especially in the $\mathrm{WebNLG}$-$\mathrm{R}$ dataset, where the performance is decreased about 8.7\% in F1 score. The result indicates that word-word interactions significantly bolster the ability of model generalization. Comparing the performance between $\mathrm{NYT}$-$\mathrm{R}$ and $\mathrm{WebNLG}$-$\mathrm{R}$, the word-word interaction is more influential in the latter. The marked difference across the two datasets stems from the greater number of relation types in $\mathrm{WebNLG}$-$\mathrm{R}$ (171 relations) relative to $\mathrm{NYT}$-$\mathrm{R}$ (24 relations). This difference highlights the critical role of word-to-word interactions in a dataset with a greater number of relations.

(2) \emph{w/o ERI} denotes the model where entity-relation interaction is removed from interaction network. Experimental findings indicate that our model's F1 score lags 0.5\% behind that of {w/o WWI} on the $\mathrm{NYT}$-$\mathrm{R}$ dataset. It suggests the entity-relation interaction strengthens the long-distance semantic dependencies between relations and entities, effectively filtering out the semantic features related to redundant relational triples. It is worth noting that the highest F1 score is achieved on the Entire indicator, which may be due to the reason that many parameters can impact the model's ability to effectively retain information.

In contrast, the degradation on the $\mathrm{WebNLG}$-$\mathrm{R}$ dataset is relatively minor. In the Partial metric, the model's F1 score is 1.2\% lower than {w/o WWI}, highlighting the importance of entity-relation interactions in effectively handling partially seen relational triples. The performance differences between the two datasets may stem from greater relational overlapping patterns (e.g., SEO, EPO and SOO). Hence, the $\mathrm{NYT}$-$\mathrm{R}$ dataset exhibits more pronounced entity-relation interactions.

(3) \emph{w/o HTI} denotes the model where head-tail interaction is removed from the interaction network. It disables the model to encode the global semantic features of relational triples. Experimental results show a significant performance decline on the $\mathrm{NYT}$-$\mathrm{R}$ dataset, particularly in the Entire and Unseen metrics. The result highlights the importance of head-tail interaction information. This interaction enhances global semantic features associated with relational triples, aiding in the remediation of triples that may have been overlooked or omitted in earlier steps.

In contrast, the performance degradation on the $\mathrm{WebNLG}$-$\mathrm{R}$ dataset is smaller. Compared to the $\mathrm{WebNLG}$-$\mathrm{R}$ dataset, the $\mathrm{NYT}$-$\mathrm{R}$ dataset contains more complex sentences and general features longer sentences. These characteristics can contribute to the performance differences observed between the two datasets. Global semantic information proves to be more effective for handling complex and lengthy sentences.

(4) \emph{w/o ALL} is the model in which word-word, entity-relation, and head-tail interactions are excluded from the interaction network. Experimental results show a significant performance decline in the two datasets. This validates the capability of our method to handle entirely seen, partially seen, and unseen relational triples. This effectiveness stems from the efficiency of our interaction network. Furthermore, it can be concluded that leveraging any two types of semantic interaction information substantially enhances the discriminability of the model. The semantic interactions among multiple objects empower the model to recognize unseen relationship triples, offering a theoretical framework for applying natural language processing to tackle various other tasks.

(5) In the Encoder module (Section \ref{sec:encode}), sentences and relations are combined for encoding. While increasing the interaction between relations and sentences, this also introduces additional complexity to the model. To explore its impact, we isolate the relation sequence from the input sentence to model the two types of interactions separately, as represented in Table \ref{table:ablation Study} under the label ``\emph{w/o SEP}''. Specifically, the same BERT encoder is used to separately derive the sequence embeddings for the input sentence and the natural language descriptions of each relation. Then, after embedding both sentences, we concatenate them together and input them into a multi-granularity interaction network. 

Given that ``\emph{w/o SEP}'' considers entities and relations as separate input representations, the deep Transformer network is only able to model correlations within entity pairs and relation pairs independently. As shown in Table \ref{table:ablation Study}, the performance of ``\emph{w/o SEP}'' significantly decreases on both datasets compared to the Fully Model, highlighting the importance of combining sentences and relations for effective encoding. We further analyze the performance of ``\emph{w/o SEP}'' and other models across different interaction types. On the $\mathrm{NYT}$-$\mathrm{R}$ dataset, Unseen outperforms ``\emph{w/o ALL}'', while Entire shows the opposite result. This further proves that the interaction network can effectively enhance the ability to identify triples consisting of ``unseen'' entities. Moreover, since the encoder module forms the foundation for subsequent modules, the lack of interactive semantic information negatively affects the performance of later stages.

\section{Analytical Experiments}
\label{sec:analyis}

\subsection{Performance on Different Overlapping Patterns}

\begin{table*}[h]
	\small
	\caption{\label{tab:results on complex scenarios} F1-score (\%) on sentences with different overlapping patterns and different numbers of triples for the NYT and WebNLG datasets. $N$ represents the number of triples within a sentence.}
	\centering
    \begin{tabular}{c |c| ccccccccc}
		\hline
		Dataset &  Model & $\mathrm{Normal}$ & $\mathrm{EPO}$ & $\mathrm{SEO}$ & $\mathrm{SOO}$ & $\mathrm{N}=1$ & $\mathrm{~N}=2$ & $\mathrm{~N}=3$ & $\mathrm{~N}=4$ & $\mathrm{~N} \geq 5$ \\
		\hline
		\multirow{7}{*}{ $\mathrm{NYT}$ } 
		& $\mathrm{CasRel}$ & 87.3 & 92.0 & 91.4 & 77.0 & 88.2 & 90.3 & 91.9 & 94.2 & 83.7 \\
		& $\mathrm{TPLinker}$ & 90.1 & 94.0 & 93.4 & 90.1 & 90.0 & 92.8 & 93.1 & 96.1 & 90.0 \\
		& $\mathrm{SPN}$ & 90.8 & 94.1 & 94.0  & 81.8 & 90.9 & 93.4 & 94.2 & 95.5 & 90.6 \\
		& $\mathrm{PRGC}$ & $91.0$ & 94.5 & 94.0 & 90.8 & 91.1 & 93.0 & 93.5 & 95.5 & 93.0 \\
		& $\mathrm{DirectRel}$ & $\mathbf{91.7}$ & 94.8 & 90.0 & 90.0 & $\mathbf{91.7}$ & $\mathbf{94.1}$ & 93.5 & 96.3 & 92.7 \\
		& $\mathrm{BiRTE}$ & 91.4 & 94.2 & 94.7 & - & 91.5 & 93.7 & 93.9 & 95.8 & 92.1 \\
		& $\mathrm{Ours}$ & 91.1 & $\mathbf{95.3}$ & $\mathbf{95.2}$ & $\mathbf{91.6}$ & 91.1 & 93.7 & $\mathbf{95.1}$ & $\mathbf{97.0}$ & $\mathbf{94.3}$  \\
		\hline
		\multirow{7}{*}{ $\mathrm{WebNLG}$ } 
		& $\mathrm{CasRel}$ & 89.4 & 94.7 & 92.2 & 90.4 & 89.3 & 90.8 & 94.2 & 92.4 & 90.9 \\
		& $\mathrm{TPLinker}$ & 87.9 & 95.3 & 92.5 & 86.0 & 88.0 & 90.1 & 94.6 & 93.3 & 91.6 \\
		& $\mathrm{SPN}$ & 89.5 & 90.8 & 94.1  & 94.6 & 89.5 & 91.3 & 96.4 & 94.7 & 93.8 \\
		& $\mathrm{PRGC}$ & 90.4 & 95.9 & 93.6  & 94.9 & 89.9 & 91.6 & 95.0 & 94.8 & 92.8 \\
		& $\mathrm{DirectRel}$ & 92.0 & 97.1 & 94.5  & 94.6 & 91.6 & 92.2 & 96.0 & 95.0 & 94.9 \\
		& $\mathrm{BiRTE}$ & 90.1 & 94.3 & $\mathbf{95.9}$ & - & 90.2 & 92.9 & 95.7 & 94.6 & 92.0 \\
		& $\mathrm{Ours}$ & $\mathbf{92.5}$ & $\mathbf{97.7}$ & 94.9 & $\mathbf{96.3}$ & $\mathbf{92.1}$ & $\mathbf{94.3}$ & $\mathbf{96.5}$ & $\mathbf{95.3}$ & $\mathbf{95.4}$  \\
		\hline
	\end{tabular}
\end{table*}

Two relational triples may share none, one or two named entities, which leads to the overlapping problem. Table \ref{table:overlapping of datasets} also gives statistical information about overlapping patterns about the $\mathrm{NYT}$-$\mathrm{R}$ and $\mathrm{WebNLG}$-$\mathrm{R}$ datasets. Compared to the original $\mathrm{NYT}$ and $\mathrm{WebNLG}$ datasets, there are fewer triples and EPO patterns in the $\mathrm{NYT}$-$\mathrm{R}$ and $\mathrm{WebNLG}$-$\mathrm{R}$ datasets. Therefore, in this section, we conducted experiments specifically on the $\mathrm{NYT}$ and $\mathrm{WebNLG}$ datasets to evaluate our model's capability in handling diverse overlapping patterns. We chose six recent top-performing models to serve as our benchmarks. The experimental result is listed in Table \ref{tab:results on complex scenarios}.

As shown from the experimental results, our proposed model achieves the best performance in 14 of the 18 subsets of the two test sets, particularly excelling in scenarios involving numerous relational triples (N > 3) and overlapping patterns. The integration of multiple semantic information interactions proves especially advantageous in complex, lengthy sentences. However, the indicators Normal, N=1, and N=2 did not achieve the best results on the NYT dataset. Our model is specifically designed to leverage multiple semantic interactions and handle complex dependencies, which are less prevalent in simpler sentences. Our approach may not provide advantages over baseline models optimized for straightforward cases. Simpler sentences often require less sophisticated semantic modeling, which limits the full utilization of our model's capabilities in these subsets.

Another advantage of our model is that it does not depend prior exposure to all relational triples encountered during training. It effectively interacts with word-to-word interactions through extracted semantic information about entities, relations, as well as head and tail representations. Therefore, our model learns richer interaction semantic information in the above complex scenarios, so as to achieve competitive performance.

\subsection{Influence of the Word-Word Interaction}

\begin{figure}
	\centering
	\includegraphics[width=0.85\linewidth]{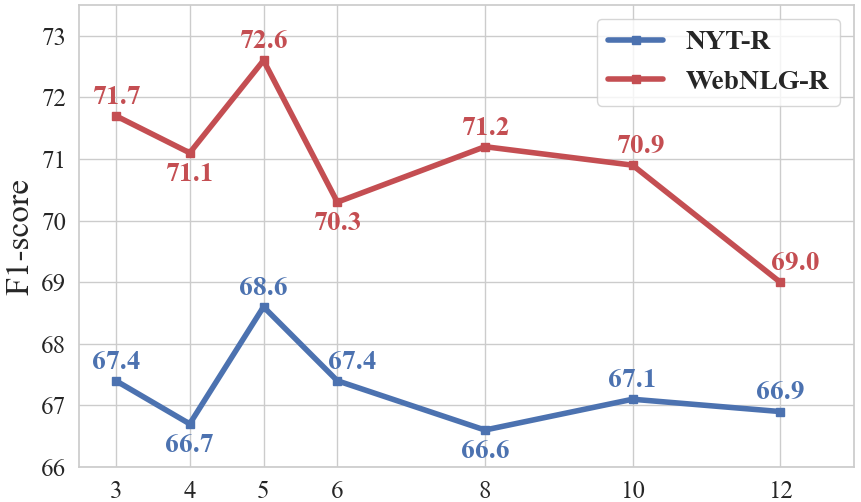}
	\caption{\label{table:effect of Word Word Interaction} The performance of tensor-based operations with different vector spaces \{3, 4, 5, 6, 8, 10, 12\}.}
\end{figure}

In our model, Word-Word Interaction is adopted to reinforce semantic features of the elements in the two-dimensional table, making them rich in relation and entity information. This experiment aims to demonstrate how the tensor-based operation affects overall performance. We evaluate our model on the $\mathrm{NYT}$-$\mathrm{R}$ and $\mathrm{WebNLG}$-$\mathrm{R}$ datasets. Table \ref{table:effect of Word Word Interaction} shows the effectiveness of tensor-based operations with vector space dimensions \{3, 4, 5, 6, 8, 10, 12\}. For a fair comparison, we trained on the $\mathrm{NYT}$-$\mathrm{R}$ and $\mathrm{WebNLG}$-$\mathrm{R}$ datasets with 20 and 50 epochs, respectively.

As shown in Table \ref{table:effect of Word Word Interaction}, performance declines when the number of tensor-based operations increases across different vector spaces, particularly at vector space dimension 12. This decline may stem from an excess of introducing vector space, which could introduce noise and potentially interfere with the model's extraction capabilities. In addition, the performance of both datasets improves when tensor-based operations across vector space 5 is introduced using the default settings of our model. This experiment underscores the value of tensor-based operations in enriching semantic information within two-dimensional tables. By leveraging interactive information between words, these operations enhance the model's memory and generalization capabilities in extracting relational triples.

\subsection{Influence of the Entity-Relation Interaction}

\begin{table}[h]
    \footnotesize
	\caption{\label{table:Influence of the Entity-Relation Interaction} The performance of CCA with different configurations on $\mathrm{NYT}$-$\mathrm{R}$ and $\mathrm{WebNLG}$-$\mathrm{R}$.}
    \centering
    \begin{tabular}{lccccccc}
        \hline 
        \multirow{2}{*}{ Model } & \multicolumn{3}{c}{ $\mathrm{NYT}$-$\mathrm{R}$ } & & \multicolumn{3}{c}{ $\mathrm{WebNLG}$-$\mathrm{R}$ } \\
        & Prec. & Rec. & F1 & & Prec. & Rec. & F1  \\
        \hline
        $CCA_{Our}(\times 1)$ & 70.2 & 67.1 & 68.6   &   & 78.1 & 67.8 & 72.6  \\
        $CCA_{Our}(\times 2)$ & 69.7 & 61.3 & 65.2   &   & 79.7 & 66.0 & 72.2  \\
        $CCA_{Ori}(\times 1)$ & 73.2 & 58.4 & 65.0   &   & 77.7 & 65.3  & 71.0  \\
        $CCA_{Ori}(\times 2)$ & 75.1 & 57.7 & 65.3   &   & 78.9 & 65.8  & 71.8  \\
        \hline
    \end{tabular}
\end{table}

In our model, the entity-relation interaction captures long-distance semantic dependencies between relations and entities, effectively filtering out the semantic features related to redundant relational triples. To assess how our optimized CCA affects ERI, we ran a set of experiments and contrasted our model's performance against the original CCA. The experiments were conducted on both $\mathrm{NYT}$-$\mathrm{R}$ and $\mathrm{WebNLG}$-$\mathrm{R}$ datasets. Table \ref{table:Influence of the Entity-Relation Interaction} summarizes these experimental results, highlighting the performance across various configurations. The notation ``Ori'' refers to the use of the convolution operation from the original CCA, while ``$\times 1/2$'' indicates that the CCA was unrolled into 1 or 2 loops.

From the experimental results presented for ``$CCA_{Our}(\times 1)$'' and ``$CCA_{Ori}(\times 1)$'', it is clear that the fully connected layer in ``$CCA_{Our}(\times 1)$'' exhibits a robust nonlinear fitting ability, which allows the model to learn task-relevant semantic features more effectively. This results in stronger performance, especially on the $\mathrm{NYT}$-$\mathrm{R}$ dataset, where ``$CCA_{Our}(\times 1)$'' outperforms ``$CCA_{Ori}(\times 1)$'' in both precision and recall.

Interestingly, for ``$CCA_{Ori}$'', unrolling the CCA into 2 loops ($\times 2$) improves the model's performance. The reason lies in the convolutional operations, which, although effective in capturing local dependencies, may struggle to model long-range semantic relationships between entities and relations. By unrolling the model into multiple loops, the convolution operation can leverage a wider context, thus better capturing these long-distance dependencies, which is in line with the findings of \cite{huang2019ccnet}.

However, for ``$CCA_{Our}$'', we observe a performance drop on the $\mathrm{NYT}$-$\mathrm{R}$ dataset when unrolled into two loops ($\times 2$). This could be due to the fact that, in this configuration, the fully connected layer already captures most of the useful semantic information when the training set is sufficiently large. When additional loops are added, the model may overfit or fail to extract additional meaningful information, leading to a reduction in performance. In contrast, when considering the $\mathrm{WebNLG}$-$\mathrm{R}$ dataset, unrolling into two loops ($\times 2$) results in a slight improvement in performance. This likely stems from the training set's limited size and the higher quantity of relation types, where extra loops help the model capture diverse semantic relationships more effectively. In such cases, the increased capacity for modeling long-range dependencies, facilitated by the unrolling of loops, can help the model navigate the complexities of the dataset more effectively.
In summary, our experiments demonstrate that the entity-relation interaction mechanism plays a significant role in improving model performance, with the fully connected layer in ``$CCA_{Our}$'' showing notable advantages in capturing semantic features. The comparison between configurations also highlights the importance of adjusting the architecture depending on the dataset size and complexity to achieve the best results.

\subsection{Influence of the Head-Tail Interaction}
\label{Influence of the Head-Tail Interaction}

\begin{table}[h]
    \small
	\caption{\label{table:Influence of the Head-Tail Interaction} The performance of Self Interaction and Multi-Object Interaction with different configurations on $\mathrm{NYT}$-$\mathrm{R}$ and $\mathrm{WebNLG}$-$\mathrm{R}$. }
    \centering
    \begin{tabular}{cccccccc}
        \hline
        \multirow{2}{*}{ Model } & \multicolumn{3}{c}{ $\mathrm{NYT}$-$\mathrm{R}$ } & & \multicolumn{3}{c}{ $\mathrm{WebNLG}$-$\mathrm{R}$ } \\
        & Prec. & Rec. & F1 & & Prec. & Rec. & F1  \\
        \hline 
        $[\mathrm{MLP}, \mathrm{CA}]$ & 71.5 & 59.7 & 65.1   &   & 78.5 & 64.0  & 70.6  \\
        $[\mathrm{MHA}, \mathrm{MLP}]$ & 73.8 & 57.4 & 64.6   &   & 79.5 & 66.4 & 72.4  \\
        $[\mathrm{MLP}, \mathrm{MLP}]$ & 73.9 & 57.0 & 64.3   &   & 78.1 & 67.8 & 72.6  \\
        $[\mathrm{MHA}, \mathrm{CA}]$ & 70.2 & 67.1 & 68.6    &   & 79.7 & 66.0 & 72.2  \\
        \hline
    \end{tabular}
\end{table}

In our model, the head-tail interaction plays a crucial role in enhancing the global semantic features associated with relational triples. This interaction helps to identify and remediate triples that may have been overlooked or omitted in earlier stages of processing, ensuring a more comprehensive understanding of the relational triples. To evaluate the impact of Self Interaction and Multi-Object Interaction on HTI, we ran a set of experiments on a pair of rearranged datasets, namely $\mathrm{NYT}$-$\mathrm{R}$ and $\mathrm{WebNLG}$-$\mathrm{R}$.

The experimental results are presented in Table \ref{table:Influence of the Head-Tail Interaction}, which compares the performance of different configurations of Self Interaction and Multi-Object Interaction within the context of HTI. Specifically, the table showcases the precision (Prec.), recall (Rec.), and F1-score for each model configuration on both datasets. The model structure ''$[\varepsilon, \delta]$'' indicates that the specified module (Self Interaction or Multi-Object Interaction) are in different configurations. For example, when we denote ``[$\mathrm{MLP}, \mathrm{MLP}$]'', it implies that both $\varepsilon$ and $\delta$ are adopted with a Multilayer Perceptron structure.

As observed from the table, the performance of different configurations varies across datasets. On the $\mathrm{NYT}$-$\mathrm{R}$ dataset, the model's performance experiences degradation to varying degrees when configured with MLP. This decline in performance highlights the indispensable role that CA and MHA play in the HTI mechanism. These attention mechanisms are far more effective than MLPs in capturing and processing the relational semantics of the entities involved in the triples. The experimental results for the configurations ``$[\mathrm{MLP}, \mathrm{CA}]$'' and ``$[\mathrm{MHA}, \mathrm{MLP}]$'' indicate a notable drop in performance, especially in recall and F1-score. This suggests that CA's capability for capturing head-tail entity interactions is essential to the HTI module's effectiveness. By attending to the semantic features of both the entities (head and tail) and the relations between them, CA enables the model to capture more nuanced, contextualized information that is crucial for accurate relational triple extraction.

Conversely, on the $\mathrm{WebNLG}$-$\mathrm{R}$ dataset, the performance of the model with ``$[\mathrm{MLP}, \mathrm{MLP}]$'' improves. This improvement may be attributed to the specific characteristics of the dataset, which contains 171 relational types and a relatively small training set. In such cases, the attention mechanisms (CA and MHA) may struggle to effectively capture the complex relations, leading to diminished performance. As a result, simplifying the model with MLP helps to enhance performance, particularly when attention mechanisms fail to generalize well in this context.

\subsection{Influence of the Number of Iterations on the Interaction Network}

\begin{table}[h]
	\caption{\label{table:iterations of the interaction network} The performance of interaction network with different number of iterations of on $\mathrm{NYT}$-$\mathrm{R}$ and $\mathrm{WebNLG}$-$\mathrm{R}$.}
    \centering
    \begin{tabular}{lccccccc}
        \hline
        \multirow{2}{*}{ Model } & \multicolumn{3}{c}{ $\mathrm{NYT}$-$\mathrm{R}$ } & & \multicolumn{3}{c}{ $\mathrm{WebNLG}$-$\mathrm{R}$ } \\
        \cline{2-4} \cline{6-8} 
        & Prec. & Rec. & F1 & & Prec. & Rec. & F1  \\
        \hline
        $I = 1$ & 70.2 & 67.1 & 68.6   &   & 78.1 & 67.8 & 72.6  \\
        $I = 2$ & 69.2 & 61.9 & 65.4   &   & 70.5 & 37.3 & 48.8  \\
        $I = 3$ & 64.6 & 64.1 & 64.4   &   & 69.4 & 36.6 & 47.9  \\
        \hline
    \end{tabular}
\end{table}

In our model, the multi-granularity interaction network has been shown to effectively refine the feature vectors for head and tail entities. To assess the number of iterations that influence model performance, we ran multiple experiments across two standard datasets, namely $\mathrm{NYT}$-$\mathrm{R}$ and $\mathrm{WebNLG}$-$\mathrm{R}$. The performance metrics, including Precision (Prec.), Recall (Rec.), and F1-score, were evaluated across different iteration settings, specifically with 1, 2, and 3 iterations of the interaction network.

Table \ref{table:iterations of the interaction network} illustrates how different iteration counts affect performance. For each dataset, we trained the models for a fixed number of epochs—10 epochs for the $\mathrm{NYT}$-$\mathrm{R}$ dataset and 20 epochs for the $\mathrm{WebNLG}$-$\mathrm{R}$ dataset—to ensure a fair comparison across different iteration configurations.

From the table, we observe that: (1) For $\mathrm{NYT}$-$\mathrm{R}$, the optimal performance occurs at 1 iteration, yielding an F1 score of 68.6. At 2 iterations, both Precision and Recall drop, resulting in a reduced F1 score of 65.4. Further increasing the iterations to 3 results in a notable drop, with the F1 score falling to 64.4. (2) For $\mathrm{WebNLG}$-$\mathrm{R}$, the pattern is somewhat similar, with the first iteration yielding the highest F1 score of 72.1. However, as the iterations increase, the performance degrades sharply, especially in the case of Recall, which drops significantly from 68.0 at 1 iteration to just 37.3 at 2 iterations. This suggests diminishing returns and potential overfitting with additional iterations.

These results indicate that while increasing the number of iterations can refine entity representations, there is a point beyond which additional iterations may lead to diminishing performance or even overfitting, particularly with more complex datasets like $\mathrm{WebNLG}$-$\mathrm{R}$. Thus, the number of iterations should be chosen carefully, balancing the benefits of interaction refinement with the potential for performance decline. In other words, when the individual models are sufficiently strong, it is possible to accurately extract the corresponding relational triples in just one iteration.

\subsection{Performance on Widely-Used Datasets}

\begin{table}[h]
    \small
    \centering
	\caption{\label{performance on traditional datasets} Performance on $\mathrm{NYT}$ and $\mathrm{WebNLG}$.}
    \begin{tabular}{lccccccc}
        \hline
        \multirow{2}{*}{ Model } & \multicolumn{3}{c}{ $\mathrm{NYT}$ } & & \multicolumn{3}{c}{ $\mathrm{WebNLG}$ } \\
        \cline{2-4} \cline{6-8} 
        & Prec. & Rec. & F1 & & Prec. & Rec. & F1  \\
        \hline
        BiRTE  & 92.2 & $\underline{93.8}$ & 93.0 &  & 93.2 & 94.0 & 93.6 \\
        OneRel  & 92.8& 92.9& 92.8     &     & 94.1& 94.4 & 94.3 \\
        BitCoin  &92.9 & 92.8 &93.1 & &$\underline{94.4}$ & $\underline{94.5}$ & 94.4 \\
        SOIRP  & $\underline{93.3}$& 92.9 & 93.1     &    & 94.3 & 94.1 & 94.2 \\
        RTF  & 93.2 & 93.4 & 93.3 &  & 94.0 & $\mathbf{95.3}$ & $\mathbf{94.6}$ \\
        BiSDRTE & 92.8 & $\mathbf{94.2}$ & $\underline{93.5}$ & & 94.1 & 94.6 & 94.4 \\
        \hline
        Ours & $\mathbf{93.5}$ & 93.7 & $\mathbf{93.6}$     &    & $\mathbf{95.0}$ & 94.0  & $\underline{94.5}$  \\
        \hline
    \end{tabular}
\end{table}

In this experiment, for a fair benchmark against additional baselines, we performed evaluations on the popular $\mathrm{NYT}$ and $\mathrm{WebNLG}$ datasets. These datasets contain primarily Entire type relational triples, along with a smaller proportion of Unseen type triples. This composition allows the evaluation to better reflect the model's memory capabilities. Table \ref{performance on traditional datasets} presents the outcomes.

In Table \ref{performance on traditional datasets}, our model is compared with six strong baselines: BiRTE \citep{ren2022simple}, OneRel \citep{shang2022onerel}, BitCoin \citep{he2023bitcoin}, SOIRP \citep{dai2024soirp}, RTF \citep{an2024rtf} and BiSDRTE \citep{chen2024comprehensive}. Our model exhibits strong performance, as evidenced by the experimental results. It achieves similar performance across the $\mathrm{NYT}$ and $\mathrm{WebNLG}$ datasets, outperforming almost all of compared baselines. These findings suggest that the interaction network positively impacts the model's memory capacity, driven by the integration of word-word, entity-relation, and head-tail semantic interactions.

\subsection{Computational Efficiency}

\begin{table}[h]
	\caption{\label{table:computational efficiency} Comparison of model efficiency.}
    \centering
	\begin{tabular}{l ccc}
		\hline
		\multirow{2}{*}{Model} & \multicolumn{3}{c}{$\mathrm{NYT}$-$\mathrm{R}$} \\
		& Training Time & GPU Memory & F1  \\
		\hline
		$\mathrm{SOIRP}$ & 6738 seconds & 4278 MB & 67.5 \\
		\hline 
		${Ours}$ & 7113 seconds & 5180 MB & 68.6 \\
		\quad ${Ours}_{w/o \mathrm{WWI}}$ & 6474 seconds & 4612 MB & 67.6 \\
		\quad ${Ours}_{w/o \mathrm{ERI}}$ & 5724 seconds & 4338 MB & 67.1 \\
		\quad ${Ours}_{w/o \mathrm{HTI}}$ & 6418 seconds & 5042 MB & 67.2 \\
		\quad ${Ours}_{w/o \mathrm{ALL}}$ & 4779 seconds & 3144 MB & 64.6 \\
		\quad ${Ours}_{w/o \mathrm{SEP}}$ & 9363 seconds & 5170 MB & 62.8 \\
		\hline
	\end{tabular}
\end{table}

In this section, we present an analysis of model efficiency measured by training time and GPU memory usage. We tested on the $\mathrm{NYT}$-$\mathrm{R}$ dataset with a batch size of 1. A single NVIDIA RTX A6000 GPU, with default configuration, was used for all experiments. The outcomes are reported in Table  \ref{table:computational efficiency}, which includes the Training Time (in seconds), GPU Memory usage (in MB), and F1-score for each model.

As shown in the table, our model achieves a significant F1-score gain over the previous method (SOIRP), albeit with increased memory usage and somewhat extended training time. Specifically, Ours takes 7113 seconds per epoch to train, occupying 5180 MB of GPU memory, and achieves an F1-score of 68.6. This performance surpasses that of SOIRP, which has an F1-score of 67.5, although it requires less memory (4278 MB) and takes slightly less time (6738 seconds).

Furthermore, we report the outcomes of ablation experiments in which various modules of our framework were removed to examine their effects on efficiency and performance. Notably, ``${Ours}_{w/o \mathrm{SEP}}$'' results in a 30\% increase in training time compared to our approach. This indicates that separately encoding relations and sentences not only reduces performance but also significantly increases training time. While WWI, ERI, and HTI increase both training time and GPU memory usage, they contribute to improved performance.

Overall, the results highlight the trade-offs between memory usage, training time, and performance. While the inclusion of each component of our model increases memory consumption and training time, it leads to improved model performance. Lower memory usage would likely boost our approach's performance.

\subsection{Case Study}

\begin{figure*}[h] 
	\centering
	\includegraphics[width=0.96\linewidth]{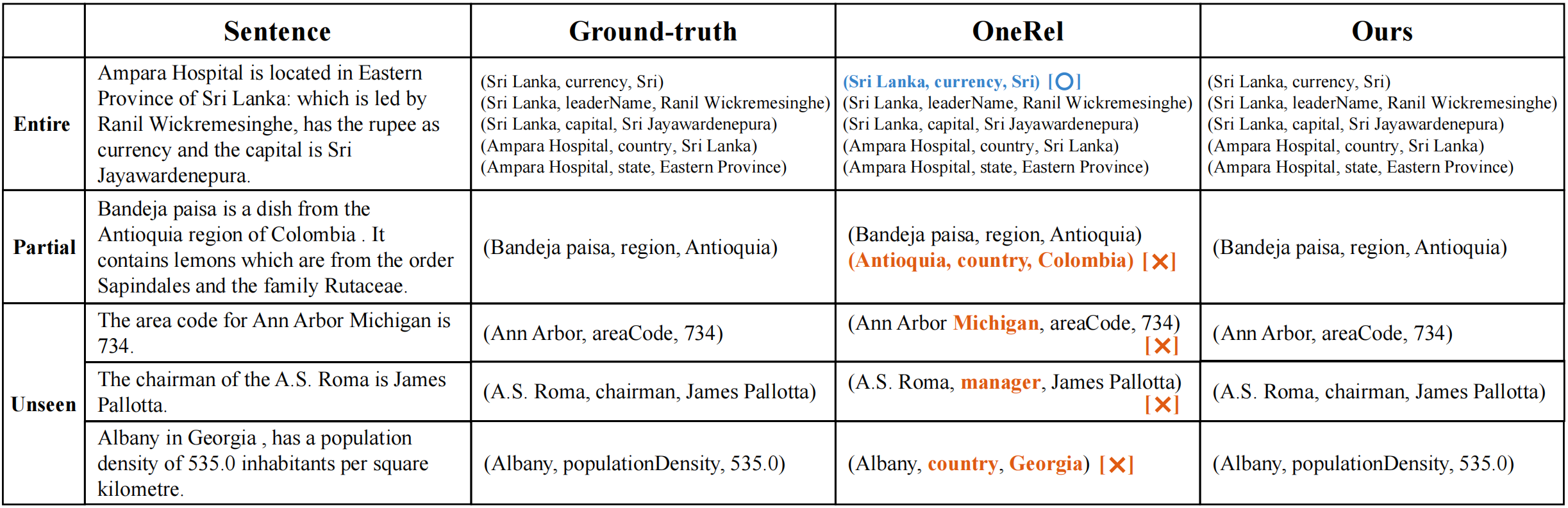}
	\caption{\label{table:case study} Examples of sentences of the Entire, Partial, and Unseen three triple types from OneRel and our models and the corresponding predictions. The orange mark indicates an error in the extracted relational triples. The blue mark indicates that the extracted relational triple is missing.}
\end{figure*}

Figure \ref{table:case study} presents a comparison between OneRel and our model across five examples sourced from the WebNLG-R dataset. These examples can be divided into three triple types: Entire, Partial, and Unseen. For the Entire, we selected five sentences containing relational triples (including $SOO$ and $SEO$ patterns) to evaluate memory capability. For the Partial, we chose a long and complex sentence containing one relational triple. For the Unseen, three short sentences were selected, each containing one relational triple that was unseen during training.

In the entirely seen examples, our model outperforms OneRel, which fails to predict the relational triple (Sri Lanka, currency, Sri) of the $SOO$ pattern. In partially seen examples, OneRel often predicts redundant relational triples prone to inaccuracies due to noise. In unseen examples, where relational triples are entirely new, OneRel exhibits errors such as insensitivity to unseen entity spans and misjudgment of relation types between entity pairs.

Our model leverages comprehensive semantic interaction information among multiple objects, whereas OneRel only captures partial semantic information, resulting in inferior predictions. These observations underscore the effectiveness of integrating multiple semantic interactions for relational triple extraction, enhancing both memory and generalization capabilities.

\subsection{Visualization Of Different Model Components}

In this visualization, we focus on depicting Word-Word and Entity-Relation Interactions, excluding Head-Tail Interactions. The Head-Tail Interaction component transforms the two-dimensional sentence representation into a one-dimensional sequential form, integrating the global semantic aspects of the sentence. One-dimensional sequence formats are often difficult to visualize. We gather two sentences from the $\mathrm{NYT}$-$\mathrm{R}$ dataset, which encompass various types of relational triples. Below is an introduction to these collected sentences:

1) 1-th sentence: the sentence states, ``Passed away surrounded by her family on Friday, July 22, 2005, in Regina, Saskatchewan, Canada, at age 82''. It contains five relational triples as: (Canada, contains, Saskatchewan), (Canada, contains, Regina), (Canada, administrative, Saskatchewan), (Sa-skatchewan, country, Canada) and (Saskatchewan, contains, Regina). Partially seen relational triples include: (Canada, contains, $\cdot$), ($\cdot$, country, Canada) and (Canada, administrative, $\cdot$). Entirely seen relational triples are: (Saskatchewan, contains, Regina) and (Canada, contains, Regina). In addition, it incorporates EPO, SEO and SOO overlapping patterns.

2) 2-th sentence: the sentence states, ``Albert Scott Crossfield was born in Berkeley, Calif., on Oct. 2, 1921''. It contains one Unseen relational triples as: (Albert Scott Crossfield, birthplace, Berkeley), and it does not have any overlapping patterns.

\begin{figure}
	\centering
	\includegraphics[width=0.85\linewidth]{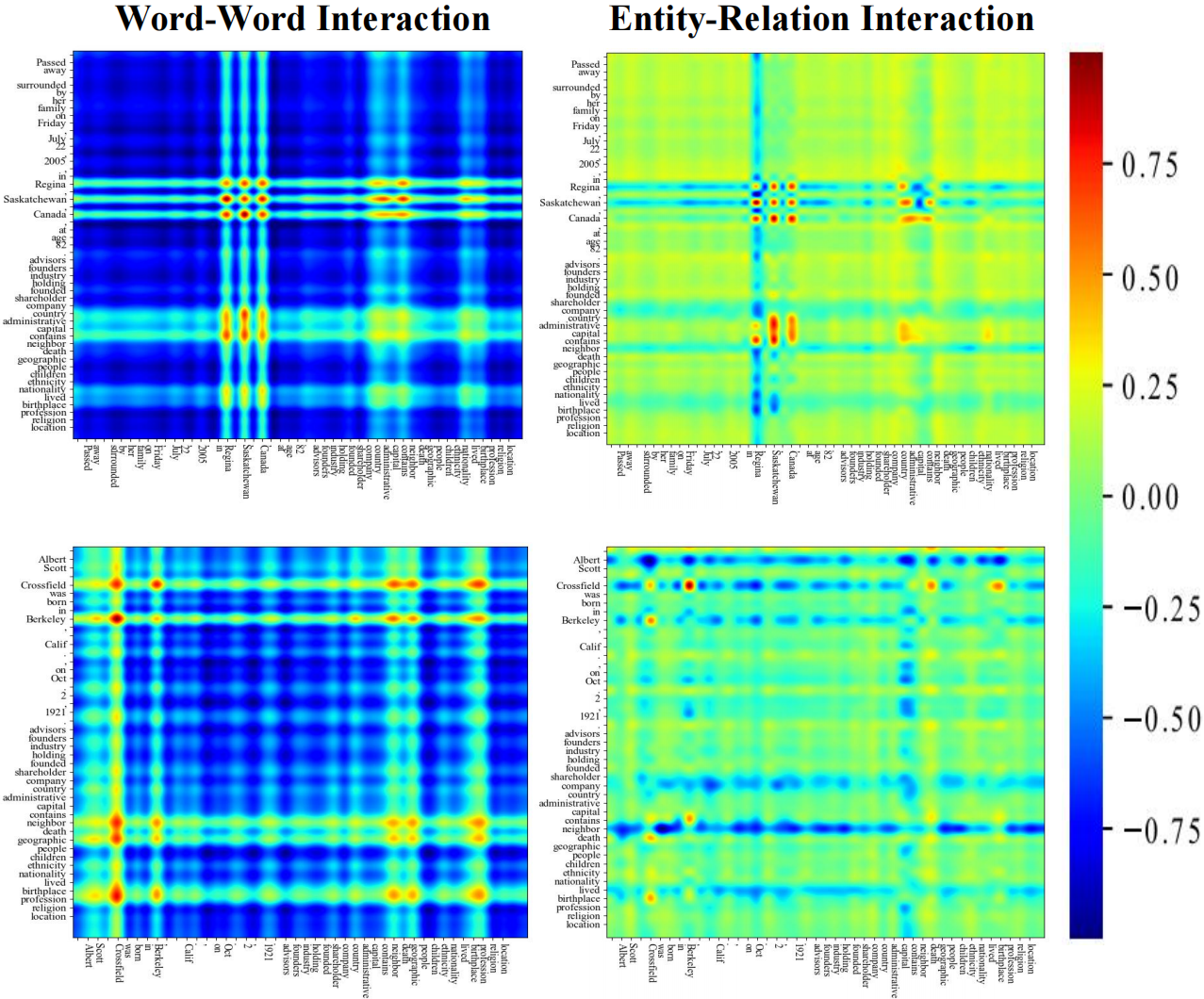}
	\caption{\label{fig:visualization} Visualization of different model components for two sentences.}
\end{figure}

For visual analysis, each 2D representation($M^{er}, \hat{M}^{er} \in \mathbb{R}^{N \times N \times d_r}$) is converted into an image($M^{er}, \hat{M}^{er} \in \mathbb{R}^{N \times N}$) using a global average pooling operation. The resulting image functions as a heat map that reveals the semantic plane. Visualizations of these semantic maps for the two example sentences are provided in Figure \ref{fig:visualization}, where color intensity reflects the semantic strength of each element within the plane.

It is noteworthy that following the utilization of Word-Word Interaction, certain important semantic information exhibits distinct contour features, although there are instances of noise between entities and relations. For instance, in 1-th sentence, entities like (Canada, Canada), (Regina, Regina), and (Saskatchewan, Saskatchewan) demonstrate stronger semantic strength. The semantic relations within entity, Head-Relation, and Relation-Tail pairs have also been moderately enhanced. Consequently, redundant semantic features in Head-Relation and Relation-Tail pairs, such as 'nationality,' 'lived,' and 'birthplace' relations, may emerge. This could be attributed to the insensitivity of word-to-word interactions towards specific entity pairs and relations.

Subsequently, after applying Entity-Relation interaction, these redundant relational semantic features are attenuated while preserving all pertinent semantic information within the sentence. For example, in 1-th sentence, redundant relational semantic features like 'nationality,' 'lived,' and 'birthplace,' which were generated in the previous stage, diminish and disappear. Meanwhile, the correct entity pairs, head-relation, and relation-tail pairs are retained. This observation is further validated in the case of ``Unseen''. This effectiveness comes from the way semantic information flows between entities and relations, which helps identify and retain useful semantic patterns while cutting down on relation redundancy.

\section{Conclusion}
\label{sec:conclusion}

This paper introduces a multi-granularity interaction network designed for jointly extracting \emph{entirely seen}, \emph{partially seen}, and \emph{unseen} relational triples. Our model leverages the semantic interactions of word-word interaction, entity-relation interaction and head-tail interaction. The word-word interaction reinforces the semantic features of elements in a 2D sentence encoding, which makes the use of relation and entity information. The entity-relation interaction captures long-distance semantic dependencies between relations and entities, effectively filtering out the semantic features related to redundant relational triples. The head-tail interaction enhances the global semantic features associated with relational triples, aiding in the remediation of triples that may have been overlooked or omitted in earlier steps. By incorporating multiple iterations, the multi-granularity interaction network is effective to progressively refine the representations of head and tail entities.
After evaluation on two rearranged datasets, the experimental results show clear and consistent improvements over the compared baselines, and establish the capability of our approach in relational triple extraction. We perform detailed analyses to understand how each model component affects performance and to confirm its importance.
Future research efforts may develop this model into a framework capable of supporting a spectrum of related multi-objective natural language processing tasks.

\section*{Limitations}

The table-filling method involves creating a two-dimensional representation of sentences. This approach not only effectively captures the semantic structure of the sentence but also allows for visualizing the model's components, thereby enhancing interpretability. However, it comes with a computational complexity of $O(N^2)$, which is notably larger compared to the $O(N)$ complexity of sequence representation. Consequently, the table-filling method requires more memory than other approaches, posing a challenge for our specific method.

\section*{Acknowledgements}
The authors thank the anonymous reviewers for their valuable suggestions. This work is supported by National Key R\&D Program of China under Grant No. 2023YFC3304500 and the Major Science and Technology Projects of Guizhou Province under Grant [2024]003.

\printcredits

\bibliographystyle{cas-model2-names}
\bibliography{cas-refs}

\end{document}